\documentclass[10pt,twocolumn,letterpaper]{article}
\usepackage{amsmath,amsfonts,bm}

\def\eqref#1{equation~\ref{#1}}
\def\1{\bm{1}}

\def\vdelta{{\bm{\delta}}}

\def\vr{{\bm{r}}}

\def\vw{{\bm{w}}}
\def\vx{{\bm{x}}}

\def\vH{{\bm{H}}}
\def\vL{{\bm{L}}}

\DeclareMathAlphabet{\mathsfit}{\encodingdefault}{\sfdefault}{m}{sl}
\SetMathAlphabet{\mathsfit}{bold}{\encodingdefault}{\sfdefault}{bx}{n}

\DeclareMathOperator*{\argmin}{arg\,min}

\usepackage{cvpr}              % To produce the CAMERA-READY version
\usepackage{graphicx}
\usepackage{amsmath}
\usepackage{amssymb}
\usepackage{booktabs}
\usepackage{array}
\usepackage[table]{xcolor}

\usepackage{makecell}
\usepackage{multirow}
\graphicspath{{Figures/}}
\usepackage{algorithm}
\usepackage{algorithmic}
\usepackage{threeparttable}
\usepackage[accsupp]{axessibility}
\usepackage[pagebackref,breaklinks,colorlinks]{hyperref}
\newcommand*\samethanks[1][\value{footnote}]{\footnotemark[#1]}

\usepackage[capitalize]{cleveref}
\crefname{section}{Sec.}{Secs.}
\Crefname{section}{Section}{Sections}
\Crefname{table}{Table}{Tables}
\crefname{table}{Tab.}{Tabs.}

\begin{document}

\title{Frequency-driven Imperceptible Adversarial Attack on Semantic Similarity}
\author{Cheng Luo\thanks{Equal Contribution} ~~~
Qinliang Lin\samethanks[1] ~~~ 
Weicheng Xie\thanks{Corresponding Author} ~~~ 
Bizhu Wu ~~~ 
Jinheng Xie ~~~
Linlin Shen\\
\textsuperscript{\rm1}Computer Vision Institute, School of Computer Science $\&$ Software Engineering, Shenzhen University\\
\textsuperscript{\rm2}Shenzhen Institute of Artificial Intelligence $\&$ Robotics for Society\\
\textsuperscript{\rm3}Guangdong Key Laboratory of Intelligent Information Processing \\
{\tt\small \{luocheng2020, linqinliang2021\}@email.szu.edu.cn,}
{\tt\small \{wcxie, llshen\}@szu.edu.cn}
}
\maketitle

%%%%%%%%% ABSTRACT
\begin{abstract}
  Current adversarial attack research reveals the vulnerability of learning-based classifiers against carefully crafted perturbations. 
  However, most existing attack methods have inherent limitations in cross-dataset generalization as they rely on a classification layer with a closed set of categories.
  Furthermore, the perturbations generated by these methods may appear in regions easily perceptible to the human visual system (HVS).
  To circumvent the former problem, we propose a novel algorithm that attacks semantic similarity on feature representations.
  In this way, we are able to fool classifiers without limiting attacks to a specific dataset.
  For imperceptibility, we introduce the low-frequency constraint to limit perturbations within high-frequency components, ensuring perceptual similarity between adversarial examples and originals.
  Extensive experiments on three datasets (CIFAR-10, CIFAR-100, and ImageNet-1K) and three public online platforms indicate that our attack can yield misleading and transferable adversarial examples across architectures and datasets.
  Additionally, visualization results and quantitative performance (in terms of four different metrics) show that the proposed algorithm generates more imperceptible perturbations than the state-of-the-art methods.
  Code is made available at \url{https://github.com/LinQinLiang/SSAH-adversarial-attack}.

\end{abstract}

%%%%%%%%% BODY TEXT
\section{Introduction}
\label{sec:intro}

With the advent of deep learning, neural network models \cite{krizhevsky2012imagenet,simonyan2014very,he2016deep,dense-net} have demonstrated revolutionary performance in recognition tasks of real-world datasets.
Nevertheless, the vulnerability of deep neural networks (DNNs) to image corruptions and adversarial examples has been unveiled \cite{szegedy2014intriguing,goodfellow2014explaining}. 
\begin{figure}[h!]
    \centering
    \includegraphics[width=0.98\columnwidth]{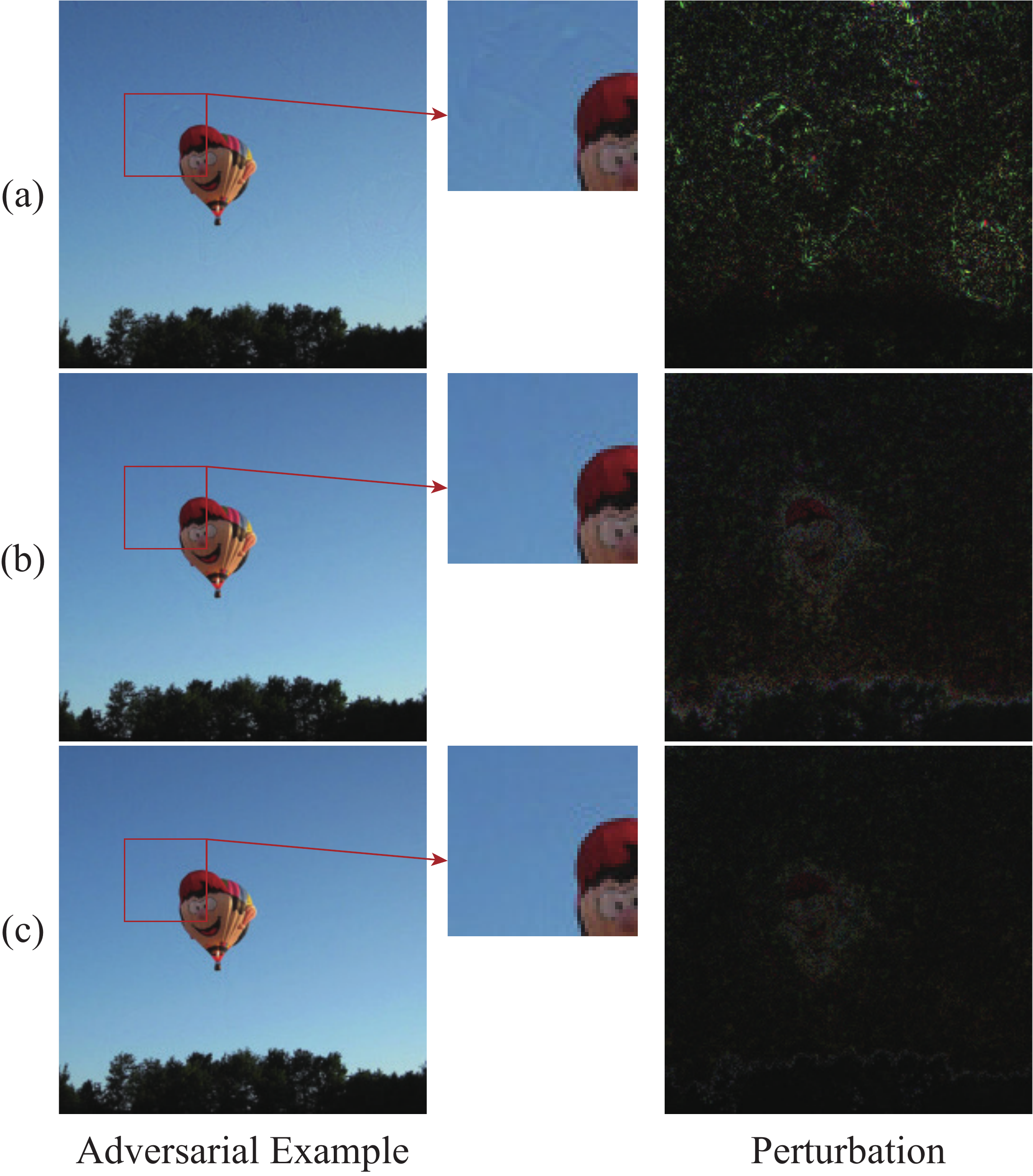} 
        \caption{Comparison of the adversarial examples and perturbations generated by three different attack methods: (a) C$\&$W, (b) Our SSA (semantic similarity attack), and (c) Our SSAH (semantic similarity attack on high-frequency components). 
        For the visualization, we regularize the perturbation by taking its absolute value and multiplying it by 25.
        }
    \label{fig:intro}
    \vspace{-0.2in}
\end{figure}
This problem hinders the applications of DNNs in security-critical domains and promotes research on understanding the robustness of DNNs, including adversarial attack \cite{goodfellow2014explaining, carlini2017towards} and defense \cite{madry2017towards, zhang2019theoretically,TramerKPGBM18,wong2018provable}.

The most intuitive approaches for white-box attacking are to increase the cost of the classification loss \cite{goodfellow2014explaining} to yield adversarial examples via gradient descent. 
Besides, they further apply $\ell_p$ distance to constrain the visual differences between benign and perturbed images.
However, conventional approaches may suffer from the two open problems:

\begin {itemize}
\item \textbf{Inherent limitation in cross-dataset generalization.}
Due to the classification layer with learned weight vectors representing specific class proxies,  current attack paradigms based on a white-box or surrogate classifier are limited to this setting, where images of the model training and attack domains are from the same set of categories. 
In real-world scenarios, however, an image from an open set \cite{panareda2017open} may belong to an unknown category to the classifier.

\item \textbf{Poor imperceptibility to HVS.} 
Sharif \etal~\cite{Sharif_2018_CVPR_Workshops} have demonstrated that the $\ell_{p}$ distance metric is insufficient for assessing perceptual similarity.
In other words, visual imperceptibility may not be explicitly reflected using only the perturbation intensity.
For instance, C$\&$W \cite{carlini2017towards}, a well-known attack method, generates easy-to-perceive perturbations on the smooth background, as shown in \cref{fig:intro}~(a).
\end {itemize}

Intuitively, a natural approach to circumvent the classification layer is to perform attacks in the feature space.
In this work, we propose a general adversarial attack, namely semantic similarity attack (SSA), which builds on the similarity of feature representations.
More specifically, we push apart the representations of adversarial and benign examples but pull that of adversarial and target (the most dissimilar) examples together.
In this way, we can fool classifiers without the knowledge of the specific image category.
The underlying assumption is that the high-level representation implies image discrimination and semantics. 
Hence, perturbing such representation can guide perturbations towards semantic regions within pixel space.
As shown in \cref{fig:intro} (b), SSA focuses on perturbing semantic regions such as objects in the scene while suppressing redundant perturbations on irrelevant regions.

In addition to $\ell_{p}$ norms \cite{carlini2017towards,chen2018ead,papernot2016limitations,rony2019decoupling}, other measures such as CIEDE2000 \cite{zhao2020towards}, SSIM \cite{hameed2021perceptually} and LPIPS \cite{laidlaw2020perceptual} are applied to approximate perceptual similarity.
In this work, we provide a different metric from the frequency domain perspective.
Generally, the low-frequency component of an image contains the basic information, whereas the high-frequency components represent trivial details and noise.
Inspired by it, we measure the variations of low-frequency components as the perceptual variations in image pixel space.
We further build a low-frequency constraint to limit the perturbations within imperceptible high-frequency components.
As depicted in \cref{fig:intro} (c), the perturbations generated by the proposed framework, \emph{i.e.}, SSAH, appear mostly on imperceptible regions such as object edges.
Some works show that adversarial examples may be neither in high-frequency nor low-frequency components \cite{maiya2021frequency}, and low-frequency perturbations with much perceptibility are especially effective for attacking defended models \cite{sharma2019effectiveness}.
Nevertheless, we consider that developing attacks in high-frequency components is significant, as it helps improve perturbation imperceptibility to HVS and learn robust models that better align with human perception.
Recent works \cite{wang2020high,NEURIPS2019_b05b57f6} also prove that these high-frequency signals are barely perceivable to HVS but can largely determine the prediction results of DNNs.

The main contributions can be summarized as follows:
\begin {itemize}
\item We propose a novel adversarial attack, SSA, which is applicable in wide settings by attacking the semantic similarity of images.
\item We present a new perturbation constraint, the low-frequency constraint, into the joint optimization of SSA to limit perturbations within the imperceptible high-frequency components. 
\item We conduct extensive experiments on three datasets, \emph{i.e.}, CIFAR-10, CIFAR-100, and ImageNet-1K, and the experiment results show that our proposed attack outperforms the state-of-the-art methods by significantly imperceptible perturbations.
\item  Experimental results demonstrate that adversarial perturbations generated by our SSAH are more transferable across different architectures and datasets.

\end {itemize}

\begin{figure*}[t]
  \centering
  \includegraphics[width=2.0\columnwidth]{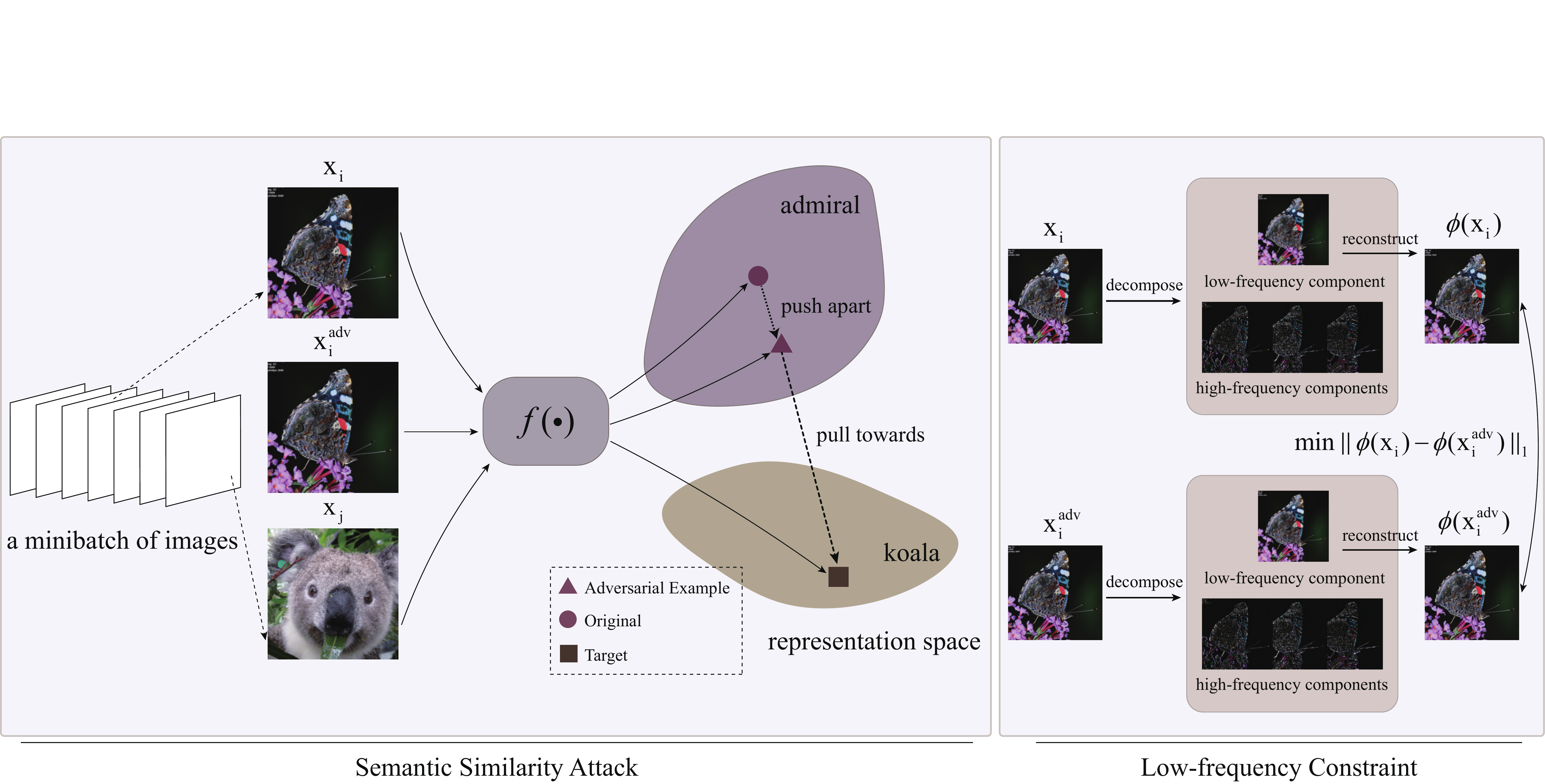}
  \caption{An overview of proposed SSAH. Left: Semantics Similarity Attack; Right: Low-frequency Constraint. $f(\cdot)$ is the mapping from an image to its embedding in representation space. $\phi(\cdot)$ is a shallow network that decomposes an image into different frequency components and reconstructs it using the low-frequency component.}
\label{fig:method_over}
\vspace{-0.1in}
\end{figure*}

\section{Related Work}
\noindent\textbf{Feature Space Attack.}
Feature space attacks \cite{sabour2016adversarial,inkawhich2019feature} manipulate the image representation to appear remarkably similar to the target image from a different class.
The same goal of these methods is to directly minimize the Euclidian distance between intermediate layer features of source and target images in the target DNN.
Similar attacks recently applied to Person Re-identification \cite{wang2020transferable, wang2019advpattern} or Image Retrieval \cite{feng2020adversarial,li2019universal} generate adversarial perturbations or patterns by minimizing the distance of the inter-class pair while maximizing the distance of the intra-class pair.
In this work, we perturb the class-specific representations of image instances from the perspective of feature similarity and design a more flexible optimization scheme.

\noindent\textbf{Imperceptible Attack.}
A rich line of works \cite{goodfellow2014explaining, carlini2017towards,hameed2021perceptually,jang2017objective,zhao2020towards,laidlaw2020perceptual, zhang2018unreasonable} resort to devising perceptual similarity metrics to constrain perturbations during adversarial example generation.
Among these metrics, $\ell_{p}$ norms of perturbations are generally employed \cite{carlini2017towards}.
However, recent works have revealed that $\ell_{p}$ norms do not well align with human perception. 
Thus, other perceptual distances in terms of similarity of object structures \cite{hameed2021perceptually}, edges \cite{jang2017objective}, color \cite{zhao2020towards}, and Learned Perceptual Image Patch Similarity (LPIPS) \cite{laidlaw2020perceptual, zhang2018unreasonable} are proposed to improve the imperceptibility of perturbations. 
In this work, we decompose images into various frequency components by wavelets and measure image pair similarity via the distance of their low-frequency components.

\noindent\textbf{Wavelets in Deep Learning.}
Wavelet is an effective tool for time-frequency analysis, and Discrete Wavelet Transform (DWT) is frequently used to decompose image data into various frequency components.
Recent works \cite{liu2018multi, duan2017sar} explore implementing wavelet transform in deep learning for various visual tasks such as image segmentation. 
In particular, Li \etal~\cite{li2020wavelet} design a DWT/IDWT layer, making discrete wavelet transform easily applicable in DNNs.

\section{Methodology}
In a white-box setup, an adversary can access details of the target classifier (\emph{i.e.}, architectures, parameters, gradients of the loss with respect to  (w.r.t.) the input) 
to craft an adversarial example $\vx^{adv} = \vx + \vdelta$ with the image perturbation $\vdelta$ to the benign example $\vx$.
Generally, a distance metric $\mathcal{D}$ is required to quantify the perceptual similarity (between an adversarial example and its original one) and is used as the constraint of the perturbation.
We can formulate the adversarial examples in the untargeted attack scenario as a solution to the following problem:
\begin{equation}\label{eq:proto}
  \begin{aligned}
      \vx^{adv} = \vx + \mathop{\arg\min}_{\vdelta} \{ \mathcal{D}(\vx, \vx + \vdelta) | \mathop{\arg\max}_{i} \{z'_i\} \neq y  \},\\
  \end{aligned}
  \end{equation} 
where $z'_i = \vw_i^T f(\vx+\vdelta)$ denotes the logit (\emph{i.e.}, the similarity score between the embedding vector $f(\vx+ \vdelta)$ of the example and the weight vector $\vw_i$ $(i= 1, 2,\ldots, C)$, $y$ denotes the ground-truth label and $C$ is the number of classes. 

In this work, we propose a novel semantic similarity attack on high-frequency components (SSAH), and the framework is depicted in \cref{fig:method_over}.
SSAH is composed of an attack paradigm (semantic similarity attack) and a new perturbation constraint (low-frequency constraint).
The semantic similarity attack does not require the classification layer but tends to change the similarity of pairwise feature representations.
The low-frequency constraint preserves the basic information of objects and limits perturbations within imperceptible high-frequency components.

\subsection{Semantics Similarity Attack}
\subsubsection{Attack Design}
Conventional white-box attack methods solve the problem presented in \cref{eq:proto} by maximizing the classification loss or changing logits.
However, given a minibatch of $N$ instances, \emph{i.e.}, $\bm{X} = [\vx_1, \vx_2,\ldots, \vx_N]$, we instead optimize the representation of the $i$-th adversarial example $\vx_i^{adv}$ as:

\newcommand{\lnorm}[1]{\lVert{#1}\rVert _2}
\begin{equation}
  \label{eq:ad_untargeted}
      \vx_i^{adv}  = \argmin_{\vx_i'} [s'_{i,i} - \min\{ s'_{i,j} | j\neq i \}]_+,
\end{equation}
where $[\cdot]_+$ denotes max$(\cdot,0)$, $\vx'_i$ is the optimization variable and initialized as $\vx_i$, $s'_{i,i} = \mathrm{sim}(f(\vx'_i),f(\vx_i))$ and $s'_{i,j} = \mathrm{sim}(f(\vx'_i),f(\vx_j))$ are similarity scores. 
In our method, we use cosine similarity of embeddings, which is defined as:

\begin{equation}
  \label{eq:cosine_similarity}
      s'_{i,j}  = \frac{f(\vx'_i)^T f(\vx_j)}{\lnorm{f(\vx'_i)}\lnorm{f(\vx_j)}}.
\end{equation}
Likewise, we can define the attack in the targeted scenario as:
\begin{equation}
    \label{eq:ad_targeted}
        \vx_i^{adv}  = \argmin_{\vx'_i} [ s'_{i,i} - s'_{i,t} ]_+,
\end{equation}
where $t$ denotes the index of the target image in the minibatch. 
\cref{eq:ad_targeted} aims to encourage the adversarial example $\vx'_i$ to be close to the target $\vx_t$ in terms of the feature representation. 
Without loss of generality, we only discuss the case of untargeted attacks.

The attack objective in \cref{eq:proto} changes logits of the classification layer. In other words, it pushes the embedding of an adversarial example apart from its ground-truth class centroid. 
By contrast, we directly change pair-wise similarity: reducing the similarity between the adversarial example and its original, while increasing the similarity between the adversarial example and its most dissimilar one in the minibatch.
In this way, our attack misleads a classifier to map the example representation into a different subspace.

\subsubsection{Self-paced Weighting}
To avoid redundant perturbations, we design a self-paced weighting scheme to improve the optimization flexibility.
This scheme design is inspired by Circle loss \cite{sun2020circle} that uses it in metric learning.
It aims at adjusting the optimization pace for each similarity score as:
\begin{equation}
      \begin{aligned}
          \vx_i^{adv}  &= \argmin_{\vx_i'} \mathcal{L}_{SSA}(\vx_i,\vx'_i) \\
          &= \argmin_{\vx_i'} [ \alpha_i s'_{i,i} - \beta_i\min\{ s'_{i,j} | j\neq i \}]_+,
      \end{aligned}
      \label{eq:ad_untargeted_alpha}
\end{equation}
where $\alpha_i$ and $\beta_i$ are adjusted in a self-paced manner as:
\begin{equation}\label{eq:untargeted_scale}
           \left\{
           \begin{aligned}
            \alpha_i&=[s'_{i,i} - m]_+,\\
            \beta_i&=[1 + m - \min\{s'_{i,j} | j\neq i \} ]_+,
           \end{aligned}
       \right.
\end{equation}
where $m \ge 0$ is a pre-defined margin. 
\cref{eq:untargeted_scale} is the weight factor setting for our attack.
Compared to the optimization of ($s'_{i,i} - s'_{i,j}$) in \cref{eq:ad_untargeted}, we introduce the adaptive weighting as ($\alpha_i s'_{i,i} - \beta_i s'_{i,j}$).
During the optimization of variable $\vx'_i$, the gradient with respect to $(\alpha_i s'_{i,i} - \beta_i s'_{i,j})$ is multiplied with $\alpha_i $ ($\beta_i$) when back-propagated to $s'_{i,i}$ ($s'_{i,j}$).
Consequently, the similarity score close to its optimum is assigned with a smaller gradient, whereas the less optimized similarity score is assigned with a larger gradient.

\subsection{Low-frequency Constraint}
\begin{figure}[htb]
    \includegraphics[width=1.0\columnwidth]{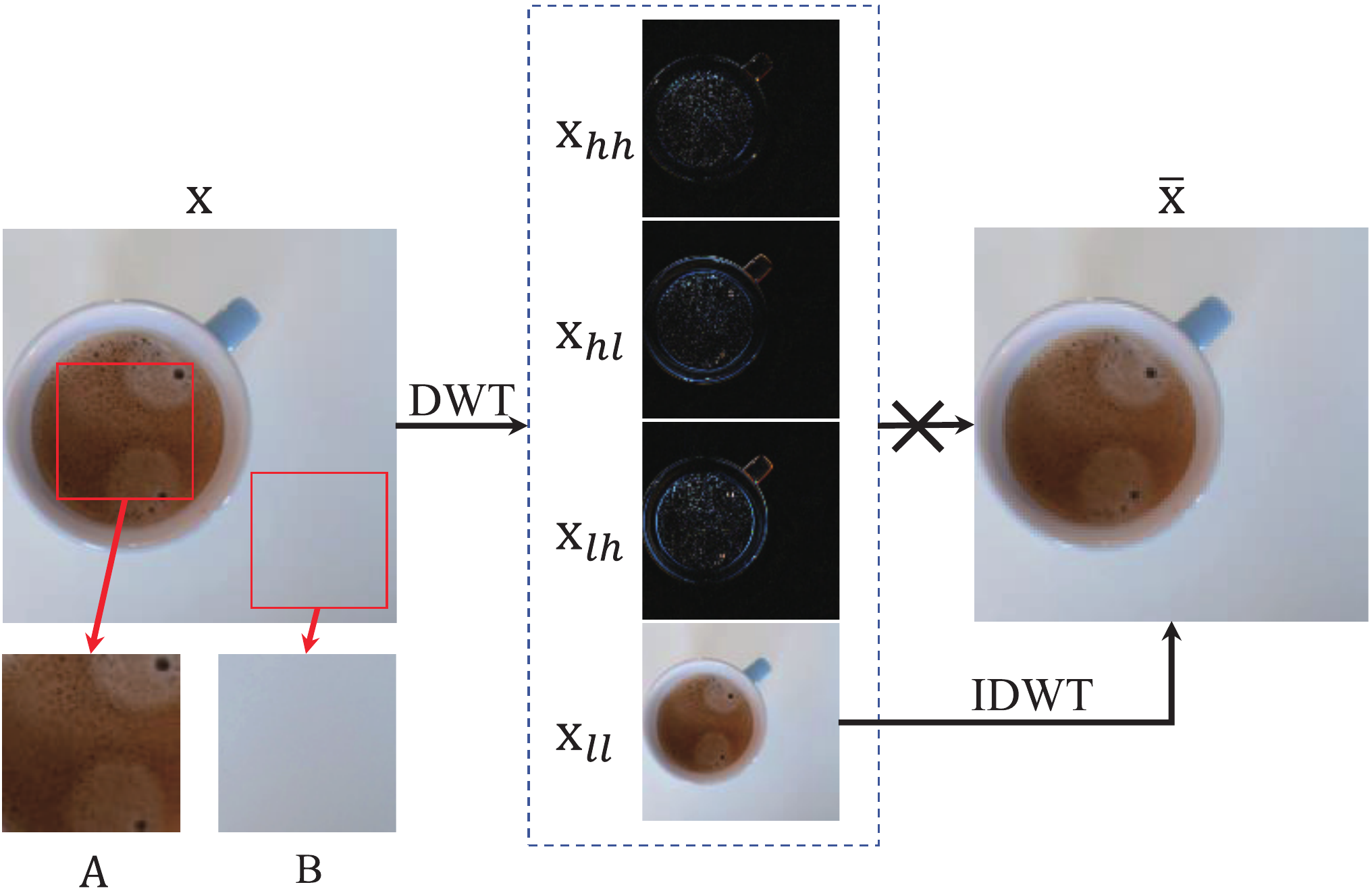}
    \caption{Illustration of our image decomposition and reconstruction by wavelet transforms. An image $\vx$ with complex (\emph{e.g.}, Part A) and smooth (\emph{e.g.}, Part B) contexts can be decomposed into the low-frequency component ($\vx_{ll}$) and high-frequency components ($\vx_{lh}$, $\vx_{hl}$ and $\vx_{hh}$) by Discrete Wavelet Transform (DWT). 
    The reconstructed image $\bar{\vx}$ has the same fundamental shape and resolution as the original image $\vx$.}
 \label{fig:high_frequency}
 \vspace{-0.1in}
\end{figure}
Although SSA yields perturbations in the representation space, there is still a risk that these perturbations may distribute in regions perceptible to HVS.
Conventional constraints, on the other hand, may result in a random distribution of perturbation.
Therefore, we seek a new constraint into the joint optimization of SSA, limiting perturbations into imperceptible details of objects.

We observe that HVS is more sensitive to object structures and smooth regions, whereas it is not easy to perceive object edges and complex textures.
For example, the perturbations hidden in the dense bubbles of Part A (in \cref{fig:high_frequency}) are more invisible than those in the smooth background like Part B.
It motivates us to limit perturbations into regions less sensitive to HVS. 

From a frequency domain perspective, the high-frequency components representing noise and textures are more imperceptible than the low-frequency component containing basic object structure. 
As a time-frequency analysis tool, discrete wavelet transform (DWT) can decompose an image $\vx$ into one low-frequency and three high-frequency components, \emph{i.e.}, $\vx_{ll}$, $\vx_{lh}$, $\vx_{hl}$, $\vx_{hh}$ as:
\begin{equation}
    \begin{aligned}
      \vx_{ll} = \vL \vx \vL^T, \vx_{lh} = \vH \vx \vL^T, \\
      \vx_{hl} = \vL \vx \vH^T, \vx_{hh} = \vH \vx \vH^T,       
    \end{aligned}
      \label{eq:dwt}
\end{equation}
where $\vL$ and $\vH$ are the low-pass and high-pass filters of an orthogonal wavelet, respectively.
As shown in \cref{fig:high_frequency}, $\vx_{ll}$ preserves the low-frequency information of the original image, whereas $\vx_{lh}$, $\vx_{hl}$ and $\vx_{hh}$ are associated with edges and drastic variations.

Normally, inverse DWT (IDWT) uses all four components to reconstruct the image. 
In this work, we drop the high-frequency components and reconstruct an image with only the low-frequency component as $\bar{\vx} = \phi (\vx)$, where
\begin{equation}
  \label{eq:idwt}
  \phi (\vx)   =  \vL^T \vx_{ll} \vL = \vL^T (\vL \vx \vL^T) \vL.
\end{equation}
Based on this process of image decomposition and reconstruction, we can obtain the main image information.
It means that we can assess the perceptual similarity between two images in terms of the main information.
On that basis, we develop a new constraint between $\vx$ and $\vx'$:
\begin{equation}
    \label{eq:constraint_D_lf}
    \mathcal{D}_{lf}(\vx,\vx')  =  \| \phi (\vx) - \phi (\vx') \|_1.
\end{equation}
Consequently, the loss of perceptual information specific to perturbed images is reduced by minimizing \cref{eq:constraint_D_lf}.

\subsection{The Unified Attack}
We define the objective of SSAH as the semantic similarity attack SSA under the new constraint $\mathcal{D}_{lf}$.
The adversarial example $\vx_i^{adv}$ can be obtained as:
\begin{equation}
    \label{eq:lossfunction_sha}
    \begin{aligned}
    \vx_i^{adv} &= \argmin_{\vx'_i} \mathcal{L}_{SSAH} (\vx_i,\vx'_i)\\
    &=  \lambda  \mathcal{D}_{lf}(\vx_i,\vx'_i) + \mathcal{L}_{SSA}(\vx_i,\vx'_i),
    \end{aligned}
\end{equation}
where $\lambda$ is a hyperparameter specific to the low-frequency constraint. In practice, we replace $\vx'_i$ with a variable $\vr_i = \mathrm{arctanh}(2\vx'_i-1)$ for optimization.
For clarity, we present the pseudo-code in \cref{al:SSAH} to outline the main procedures of our SSAH.
\begin{algorithm}[htb]
  \caption{Adversarial attack with SSAH}
  \label{al:SSAH}
  \algorithmicrequire{ A minibatch of original images $\{\vx_i\}_{i=1}^{N}$; the number of iterations $K$; the encoder $f(\cdot)$ of a classifier.}
  \begin{algorithmic}[1]
      \STATE Initialize $\{\vx'_i\}_{i=1}^{N}$ with $ \{\vx_i\}_{i=1}^{N}$;
      \FOR {$i = 1$ to $N$}
         \STATE Initialize the variable $\vr_i$ as $\mathrm{arctanh}(2\vx'_i-1)$;
          \FOR {$k = 1$ to $K$} 
              \STATE Calculate the cosine similarity scores $\{ s'_{i,j} \}_{j=1}^N$ as in \cref{eq:cosine_similarity} and use $s'_{i,i}$ and the lowest similarity score $\min\{ s'_{i,j} | j\neq i \}$ in \cref{eq:ad_untargeted_alpha};
              \STATE Calculate the constraint loss $\mathcal{D}_{lf}(\vx_i,\vx'_i)$ as in \cref{eq:constraint_D_lf};
              \STATE Optimize the variable $\vr_i$ by minimizing $\mathcal{L}_{SSAH}(\vx_i,\vx'_i)$ as in \cref{eq:lossfunction_sha} and obtain $\vx'_i$ through $\vr_i$;
          \ENDFOR
      \ENDFOR
      \RETURN $\{\vx'_i\}_{i=1}^{N}$.
  \end{algorithmic}
\end{algorithm}

\begin{table*}
  \centering
  \small
  \scalebox{1}{
  \begin{tabular}{clccccccc}
  \toprule
  Dataset & Attack & Iteration &RunTime (s) $\downarrow$ & ASR (\%) $\uparrow$ &$\ell_2$ $\downarrow$ &$\ell_{\infty}$ $\downarrow$   & FID $\downarrow$ & LF $\downarrow$\\
  \midrule[1pt] 
  \multirow{10}{*}{CIFAR-10}
      & BIM \cite{kurakin2016adversarial} & 10 & 35 & 100 &0.85 & 0.03 &14.85  & 0.23\\
      & PGD \cite{madry2017towards} & 10 & 37 & 100 &1.28 & 0.03 & 27.86 &0.34\\
      & MIM \cite{dong2018boosting}  & 10 & 46 & 100  &1.90 & 0.03 &26.00  & 0.48\\
      & AA $\ell_{\infty}$ \cite{croce2020reliable} & 100 &184 & 100 &1.91 & 0.03 &34.93 & 0.61\\ 
      & AdvDrop \cite{duan2021advdrop} & 150 & 392 & 99.92 &0.90 & 0.07 &16.34 &0.34\\
      & C$\&$W $\ell_2$ \cite{carlini2017towards} &1000 & 991 & 100  &0.39 & 0.06 &8.23 & 0.11\\
      & PerC-AL \cite{zhao2020towards} & 1000 & 1221 & 98.29 &0.86 & 0.18 &9.58  & 0.15\\
      &\cellcolor{gray!30}SSA (ours) &\cellcolor{gray!30}150 &\cellcolor{gray!30}192 &\cellcolor{gray!30}99.96  &\cellcolor{gray!30}0.29&\cellcolor{gray!30}\textbf{0.02} &\cellcolor{gray!30}5.73 &\cellcolor{gray!30}0.07\\
      &\cellcolor{gray!30}SSAH (ours) &\cellcolor{gray!30}150 &\cellcolor{gray!30}198 &\cellcolor{gray!30}99.94 &\cellcolor{gray!30}\textbf{0.26}& \cellcolor{gray!30}\textbf{0.02}&\cellcolor{gray!30}\textbf{5.03} &\cellcolor{gray!30}\textbf{0.03}\\
  \midrule[0.75pt]
  \multirow{10}{*}{CIFAR-100}
  & BIM \cite{kurakin2016adversarial} & 10 & 34 & 99.99 &0.85 & \textbf{0.03} & 15.26 & 0.32\\
  & PGD \cite{madry2017towards} & 10 & 31 & 99.99 &1.29 & \textbf{0.03} & 27.74 &0.42\\
  & MIM \cite{dong2018boosting}  & 10 & 30 & 99.99  &1.87 & \textbf{0.03} & 26.04 & 0.65\\
  & AA $\ell_{\infty}$ \cite{croce2020reliable} & 100 &184 & 100 &1.91 & \textbf{0.03} &33.86 & 0.61\\ 
  & AdvDrop \cite{duan2021advdrop} & 150 & 332 & 99.93 &0.80 & 0.07 &15.59 &0.31\\
  & C$\&$W $\ell_2$ \cite{carlini2017towards} &1000 & 751 & 100  &0.52 & 0.07 &11.04 & 0.19\\
  & PerC-AL \cite{zhao2020towards} & 1000 & 919 & 99.61 &1.41 & 0.21 &12.83  & 0.37\\
  &\cellcolor{gray!30}SSA (ours) &\cellcolor{gray!30}150 &\cellcolor{gray!30}150 &\cellcolor{gray!30}99.90  &\cellcolor{gray!30}0.48 &\cellcolor{gray!30}\textbf{0.03} &\cellcolor{gray!30}9.68 &\cellcolor{gray!30}0.17\\
  &\cellcolor{gray!30}SSAH (ours) &\cellcolor{gray!30}150 &\cellcolor{gray!30}149 &\cellcolor{gray!30}99.80 &\cellcolor{gray!30}\textbf{0.45} &\cellcolor{gray!30}\textbf{0.03}&\cellcolor{gray!30}\textbf{9.20} &\cellcolor{gray!30}\textbf{0.13}\\
  \midrule[0.75pt]
  \multirow{10}{*}{ImageNet-1K}
  & BIM \cite{kurakin2016adversarial} & 10 & 3998 &99.98 &26.85 & 0.03 & 51.92 & 11.18\\
  & PGD \cite{madry2017towards} & 10 &3451 &99.98 &54.97 & 0.03 & 45.51 &17.41\\
  & MIM \cite{dong2018boosting}  & 10 & 7847 &99.98  &91.78 & 0.03 & 101.88 & 39.42\\
  & AA $\ell_{\infty}$ \cite{croce2020reliable} & 100 &27312 & 96.97 &71.62 & 0.03 &77.49 &30.45\\ 
  & AdvDrop \cite{duan2021advdrop} & 150 &48355 & 99.76 &14.95 & 0.06 & 11.28 &5.67\\
  & C$\&$W $\ell_2$ \cite{carlini2017towards} &1000 & $>100000$ & 99.27  &\textbf{1.51} & 0.04 &12.14 & 0.67\\
  & PerC-AL \cite{zhao2020towards} & 1000 & $>100000$  & 98.78 &4.35 & 0.12 & 11.56  & 1.59\\
  & \cellcolor{gray!30}SSA (ours) &\cellcolor{gray!30}200 &\cellcolor{gray!30}35414 &\cellcolor{gray!30}98.56  &\cellcolor{gray!30}2.34 &\cellcolor{gray!30}\textbf{0.01} &\cellcolor{gray!30}4.63 & \cellcolor{gray!30}1.05\\
  & \cellcolor{gray!30}SSAH (ours) &\cellcolor{gray!30}200 &\cellcolor{gray!30}38018 &\cellcolor{gray!30}98.01 &\cellcolor{gray!30}1.81 &\cellcolor{gray!30}\textbf{0.01} &\cellcolor{gray!30}\textbf{3.90} &\cellcolor{gray!30}\textbf{0.06}\\
  \bottomrule[1pt]
  \end{tabular}} 
  \caption{Results of the attack success rate (ASR) and three metrics related with perceptual similarity by nine attack approaches in the untargeted scenario. The best results are marked in bold.}
  \label{ex:untargeted_attack}
  \vspace{-0.2in}
\end{table*}

\section{Experiments}
\subsection{Experimental Setup}
\noindent\textbf{Datasets.}
We evaluate the performance of our method on three general datasets, namely CIFAR-10 \cite{krizhevsky2009learning}, CIFAR-100 \cite{krizhevsky2009learning}, and ImageNet-1K \cite{russakovsky2015imagenet}. 
In particular, CIFAR-10 contains 50K training samples and 10K testing samples with the size of 32$\times$32 from 10 classes;
CIFAR-100 has 100 classes, containing the same number of training (testing) samples as CIFAR-10;
ImageNet-1K has 1K classes, containing about 1.3M images for training and 50K images for validation. 

\noindent\textbf{Implementation details.}
Adam optimizers with the learning rates of 0.01, 0.01, and 0.001 are used for C$\&$W, PerC-AL \cite{zhao2020towards} and our SSAH, respectively.
The default values for the hyperparameters $m$ in \cref{eq:untargeted_scale} and $\lambda$ in \cref{eq:lossfunction_sha} are 0.2 and 0.1, respectively.
The perturbation budget ($\epsilon$) is set to 8/255 under the $\ell_{\infty}$ for BIM \cite{kurakin2016adversarial}, PGD \cite{madry2017towards}, AA (AutoAttack) \cite{croce2020reliable} and MIM \cite{dong2018boosting}, respectively.
This budget is specified with the iterative step size $\alpha = 1/255$.
We use ResNet-20 models that achieve the 7.4\% and 30.4\% top-1 test errors on CIFAR-10 and CIFAR-100, respectively, as the white-box model for these two datasets.
For ImageNet-1K, pre-trained ResNet-50 that achieves the 23.85\% top-1 error is employed. 
For the DWT/IDWT layer in our low-frequency constraint, Haar wavelet is used. 
All our experiments are conducted on a NVIDIA A100 GPU with 40GB memory.

\noindent\textbf{Evaluation metrics.}
For the performance evaluation and comparison, we use the attack success rate (ASR) and four different metrics, 
including conventional average $\ell_2$ distortion, maximum perturbation intensity ($\ell_{\infty}$), Fr\'{e}chet Inception Distance (FID) \cite{heusel2017gans} and a newly introduced metric, \emph{i.e.}, average distortion of low-frequency components (LF) based on 2D DWT, for approximating the perceptual similarity. 
LF ($\mathrm{LF} = \frac{1}{N} \sum_{i=1}^N \| \phi (\vx_i) - \phi (\vx_i^{adv}) \|_2 $) is employed to quantify average variations of the basic structure information between the original and adversarial examples. 

\subsection{White-box Attacks}
In this section, we evaluate the adversarial strength and imperceptibility of the examples generated by different approaches in a white-box scenario, where the knowledge of the target system is fully accessible.

\cref{ex:untargeted_attack} shows the performances of nine attack approaches in terms of five different metrics.  
It demonstrates that our attack, with the lowest $\ell_{p}$ (\emph{i.e.}, $\ell_2$ or $\ell_{\infty}$) norm of perturbations, is successful on all three datasets.
More importantly, our semantic similarity attack (SSA), without tightly constraining the $\ell_{p}$ norms or other perceptual distances, can generate perturbations that are imperceptible.

Generally, FID is consistent with human judgment and well reflects the level of disturbance.
It calculates the distance between the benign and perturbed images in the feature space of an Inception-v3 \cite{szegedy2016rethinking} network.
The proposed SSAH achieves the FID of 3.90 on Imagenet-1K, which outperforms the state-of-the-art models like PerC-AL (11.56 FID) by a large margin.
Such improvement suggests that our feature-oriented attack generates adversarial examples with more realistic visual effects in pixel space but fewer variations in feature space.

\cref{ex:untargeted_attack} also shows that our attack significantly outperforms the other methods among all cases in terms of LF.
It implies that our attack can effectively preserve the object structure and the low-frequency component.
The results specific to the targeted attack scenario follow a similar pattern and can be found in the supplementary material.

\begin{figure}[t!]
  \centering
  \includegraphics[width=0.97\columnwidth]{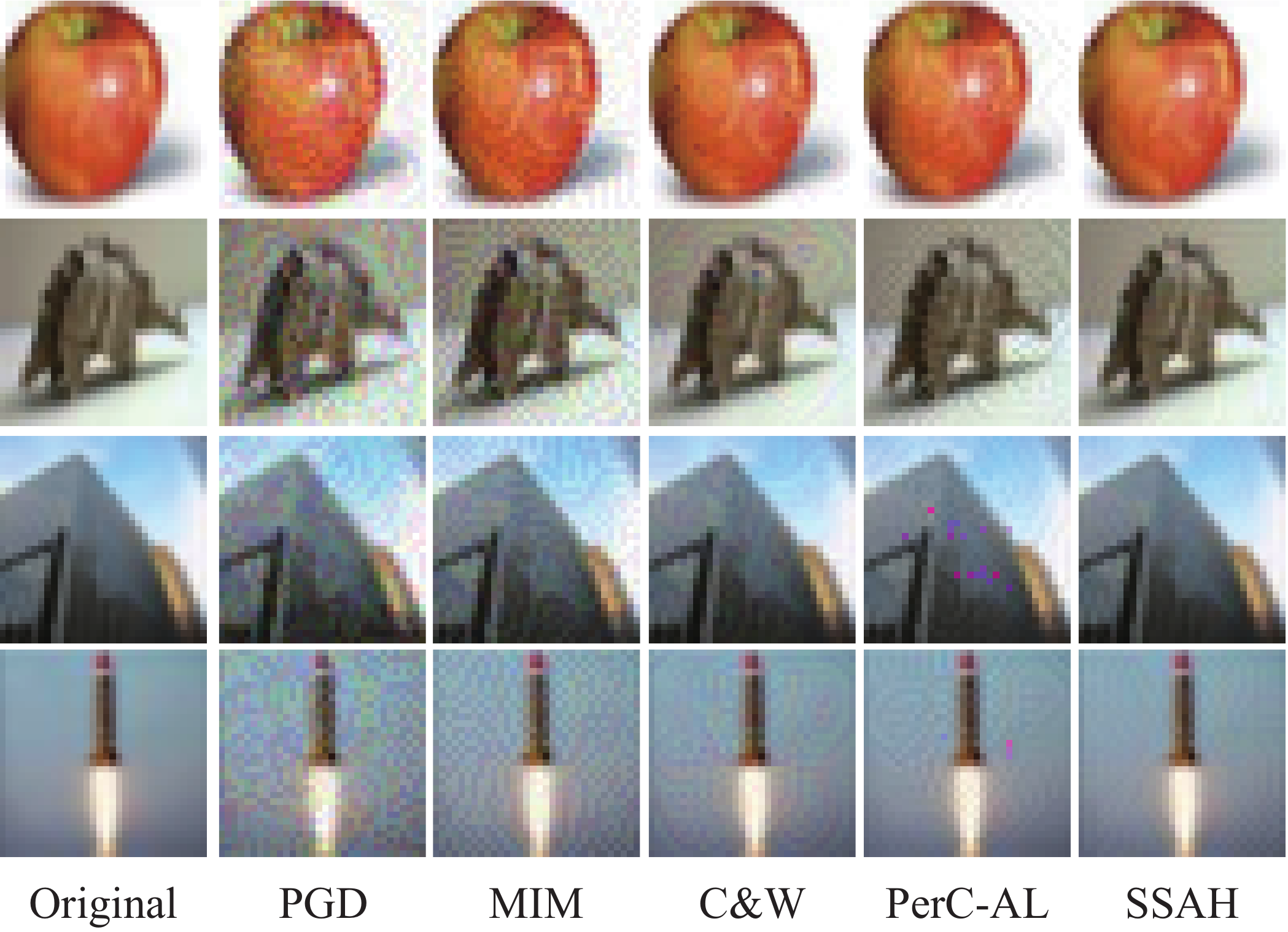}
      \caption{Adversarial examples generated by five different attack approaches on CIFAR-100. }
  \label{fig:ex_cifar}
  \vspace{-0.1in}
\end{figure}
\begin{figure*}[t]
  \centering
  \includegraphics[width=2\columnwidth]{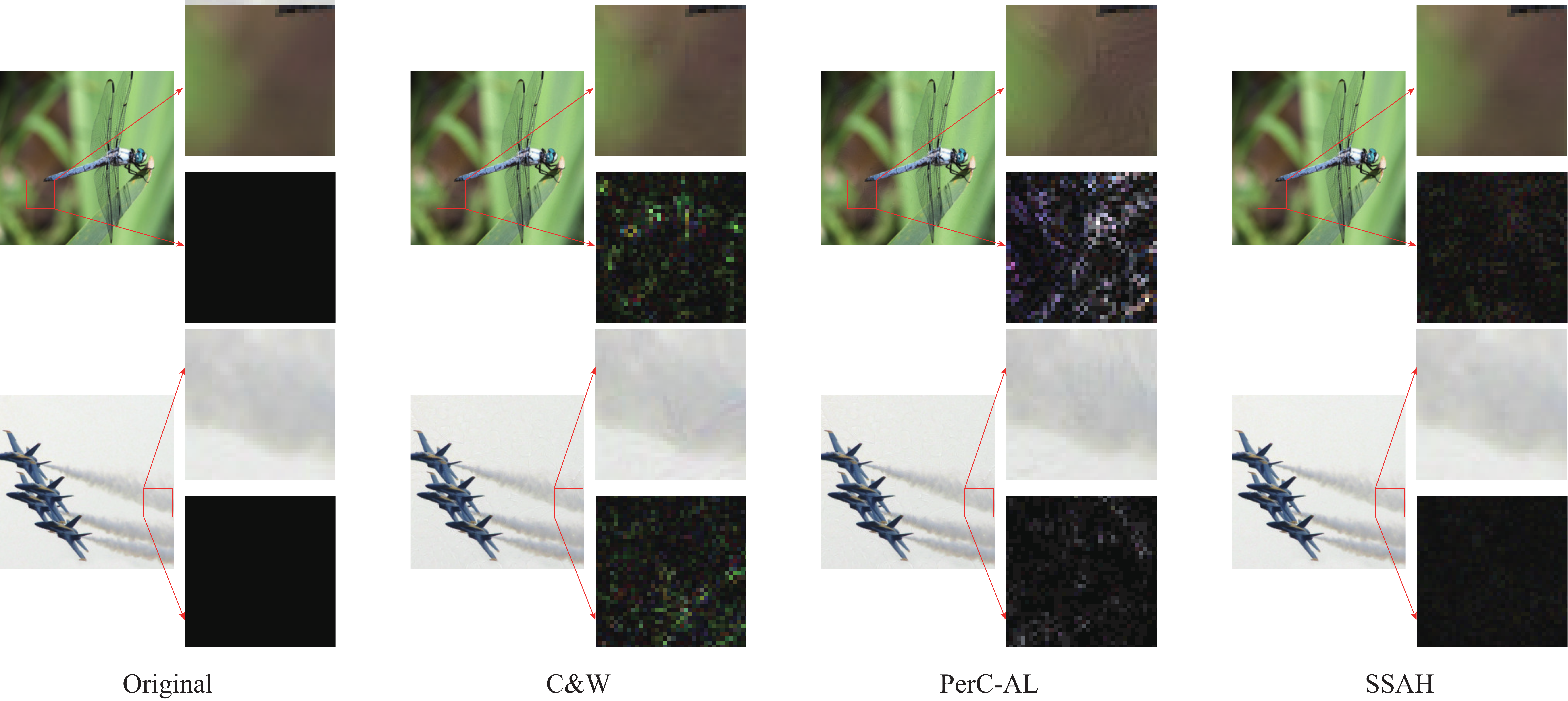} 
      \caption{Adversarial examples and perturbations generated by three attack approaches on two high-resolution images from ImageNet-1K. 
      This figure is best viewed in color/screen.
      }
  \label{fig:ex_imagenet}
  \vspace{-0.1in}

\end{figure*}

\cref{fig:ex_cifar} shows the adversarial examples generated by five approaches on CIFAR-100.
Meanwhile, \cref{fig:ex_imagenet} displays adversarial examples and perturbations (with the same regularization as \cref{fig:intro}) of higher resolution images from ImageNet-1K.
It is observed that images produced by our SSAH appear more natural to HVS.

\subsection{Robustness}
\begin{table}[htb]
  \centering
  \small
  \begin{tabular}{clcc}
  \toprule
  Defense &Attack &CIFAR-10 &CIFAR-100 \\
  \midrule
  \multirow{4}{*}{FSAT \cite{zhang2019defense}}
  &{No Attack}& 89.98 &74.11\\
  &{AdvDrop \cite{duan2021advdrop}} &70.26  &40.83    \\
  &{C$\&$W \cite{carlini2017towards}} &60.60  &25.04    \\
  &{PerC-AL \cite{zhao2020towards}} &89.80 &74.00  \\
  &{\cellcolor{gray!30}SSAH (ours)} &\cellcolor{gray!30}\textbf{60.43} &\cellcolor{gray!30}\textbf{4.85}\\
  \midrule
  \multirow{5}{*}{TRADES \cite{zhang2019theoretically}}
  &{No Attack}&84.92 &56.94\\
  &{AdvDrop \cite{duan2021advdrop}} &84.42 &56.37    \\
  &{C$\&$W \cite{carlini2017towards}} &81.24 &\textbf{48.51}    \\
  &{PerC-AL \cite{zhao2020towards}} &84.70 &56.90  \\
  &{\cellcolor{gray!30}SSAH (ours)} &\cellcolor{gray!30}\textbf{78.68}  &\cellcolor{gray!30}49.23\\
  \bottomrule
  \end{tabular}
  \caption{Recognition accuracy (\%) of two defense methods under four white-box attacks.}
  \label{tab:ex_defense}
  \vspace{-0.1in}
\end{table}
To study the robustness of the proposed attack, we compare the attack success rates of four attack approaches against two defense schemes (FSAT \cite{zhang2019defense} and TRADES \cite{zhang2019theoretically}).
The same network architecture as \cite{zhang2019theoretically}, \emph{i.e.}, WRN-34-10 introduced in \cite{BMVC2016_87}, is used to generate adversarial perturbations.

Based on the network trained with the defense method FSAT, our attack decreases the model accuracy by a large margin, \emph{i.e.}, 29.55\% on CIFAR-10 and 69.26\% on CIFAR-100,
and largely outperforms other approaches on CIFAR-100, as shown in \cref{tab:ex_defense}. 
Against the more robust defense method (\emph{i.e.}, TRADES), SSAH can still achieve competitive results. 
For CIFAR-10, an improvement of 2.56\% is achieved by SSAH compared with C$\&$W.

\subsection{Transferability}
To study the transferability of the proposed algorithm in an open-set setting, we evaluate adversarial examples transferred across both architectures and datasets.
That is, without the knowledge of the training set and architecture of a black-box model (\emph{e.g.}, ResNet-18), we study to which extent attack approaches, based on another architecture (\emph{e.g.}, ResNet-20) trained on another dataset (\emph{e.g.}, CIFAR-10), affect the classification of this black-box model on the validation set (\emph{e.g.}, ImageNet-1K).   
We use Gaussian noise and input-agnostic perturbations (\emph{i.e.}, GD-UAP \cite{mopuri2018generalizable}) as the baselines and the $\ell_{\infty}$-norm bound of 10/255 for the perturbation generation.
The experimental results in \cref{tab:ex_transferability_across_dataset_architecture} show that our SSAH significantly outperforms these two baselines for eight out of eight cases.
\begin{table}[t!]
  \centering
  \scalebox{0.8}{
  \begin{tabular}{ccccc}
  \toprule
  Surrogate &Training set &Attack &ResNet-18 &VGG-16  \\
  \midrule
  \multirow{1}{*}{-} & \multirow{1}{*}{-} &Gaussian Noise &9.18  &10.20    \\\midrule
  \multirow{2}{*}{ResNet-20} & \multirow{2}{*}{CIFAR-10} &GD-UAP \cite{mopuri2018generalizable} &14.09  &11.44    \\
  &                   &\cellcolor{gray!30}SSAH (ours)  &\cellcolor{gray!30}\textbf{17.66}  &\cellcolor{gray!30}\textbf{16.87}   \\\midrule
  \multirow{2}{*}{ResNet-20} & \multirow{2}{*}{CIFAR-100} &GD-UAP \cite{mopuri2018generalizable}  &12.72  &10.22    \\
  &                   &\cellcolor{gray!30}SSAH (ours)  &\cellcolor{gray!30}\textbf{18.31}  &\cellcolor{gray!30}\textbf{18.68}  \\\midrule
  \multirow{2}{*}{VGG-11} & \multirow{2}{*}{CIFAR-10} &GD-UAP \cite{mopuri2018generalizable} &14.63  &12.38     \\
  &                   &\cellcolor{gray!30}SSAH (ours)  &\cellcolor{gray!30}\textbf{16.92}  &\cellcolor{gray!30}\textbf{17.52}   \\\midrule
  \multirow{2}{*}{VGG-11} & \multirow{2}{*}{CIFAR-100} &GD-UAP \cite{mopuri2018generalizable} &12.93  &10.39    \\
  &                   &\cellcolor{gray!30}SSAH (ours)  &\cellcolor{gray!30}\textbf{17.92} &\cellcolor{gray!30}\textbf{19.14}   \\
  \bottomrule
  \end{tabular}}
  \caption{The attack success rates (\%) of transferring adversarial examples across different architectures and datasets. The first column (Surrogate) and the second column (Training set) represent the surrogate's architecture and training set, respectively.
  The target classifier (\emph{i.e.}, ResNet-18 or VGG-16) is trained on a different dataset (\emph{i.e.}, ImageNet-1K) and the validation set of ImageNet-1K is used for the testing.}
  \label{tab:ex_transferability_across_dataset_architecture}
  \vspace{-0.1in}
\end{table}

To test attack effectiveness in real-world scenarios, we conduct experiments of attacking the online models on Microsoft Azure\footnote{https://azure.microsoft.com/}, Tencent Cloud\footnote{https://cloud.tencent.com/} and Baidu AI Cloud\footnote{https://cloud.baidu.com/}.
The model and training data used in their platforms are completely unknown to us. 
We randomly sample 200 images from the ImageNet-1K validation set (the image names are listed in our supplementary material) and perturb them using four attack approaches on ResNet-152.
\begin{table}[h]
  \centering
  \scalebox{0.85}{
  \small
  \begin{tabular}{cccc}
  \toprule
   Attack &Microsoft Azure &Tencent Cloud &Baidu AI Cloud\\
  \midrule
  {AdvDrop \cite{duan2021advdrop}} &16.26 &15.83  &17.82 \\
  {C$\&$W \cite{carlini2017towards}} &13.82 &21.71 &27.72  \\
  {PerC-AL \cite{zhao2020towards}} &15.45 &18.61 &18.81\\\rowcolor{gray!30}
  {SSAH (ours)} &\textbf{18.70} &\textbf{37.98}  &\textbf{36.63} \\
  \bottomrule
  \end{tabular}}
  \caption{The attack success rates (\%) of transferring adversarial examples to three online models.}
  \label{tab:ex_transfer_openai}
  \vspace{-0.1in}
\end{table}
\cref{tab:ex_transfer_openai} shows the attack success rates of these attacks against the online models.
In \cref{tab:ex_transfer_openai}, the proposed attack, without any classification query, achieves an attack success rate of 37.98\% against the online model on Tencent Cloud, which outperforms the other approaches by a large margin. 

\subsection{Analysis}
In this section, we give insight into the working mechanism of the proposed attack and study the behavior of each component in the attack.

\begin{figure}[htb]
    \centering
    \includegraphics[width=1\columnwidth]{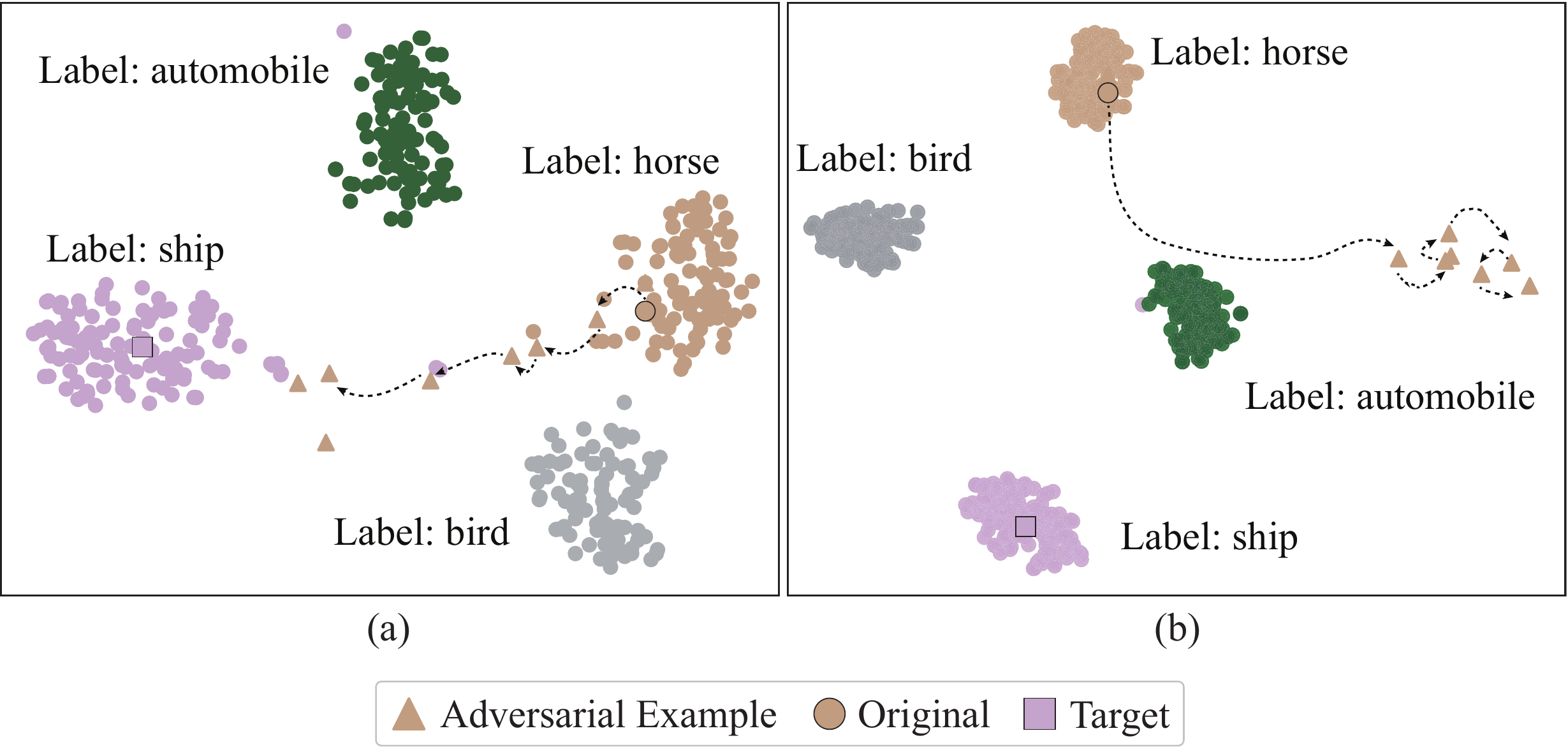}
    \caption{The 2D feature representation of the adversarial examples using the t-SNE algorithm under (a) SSAH and (b) C$\&$W. An adversarial example representation gradually updates from its original class (horse) to the selected target class (ship). The results in the iteration of 10, 15, 20, 30, and 40 are presented.}
    \label{fig:ex_semantics_attack}
    \vspace{-0.1in}
\end{figure}
To study the adversarial example generation of the proposed semantic similarity attack, we visualized the iterative adversarial examples on the 2D plane in \cref{fig:ex_semantics_attack}.
In this analysis, a subset of random instances of four classes from CIFAR-10 is used for visualization.
\cref{fig:ex_semantics_attack} shows our attack can iteratively push the adversarial example away from the benign example, and gradually guide it toward the target class in the feature representation space.
Compared with C$\&$W, our semantic similarity attack is more effective at misleading a white-box network into mapping an image to the target class subspace.

\begin{figure}[htb]
  \centering
  \includegraphics[width=1\columnwidth]{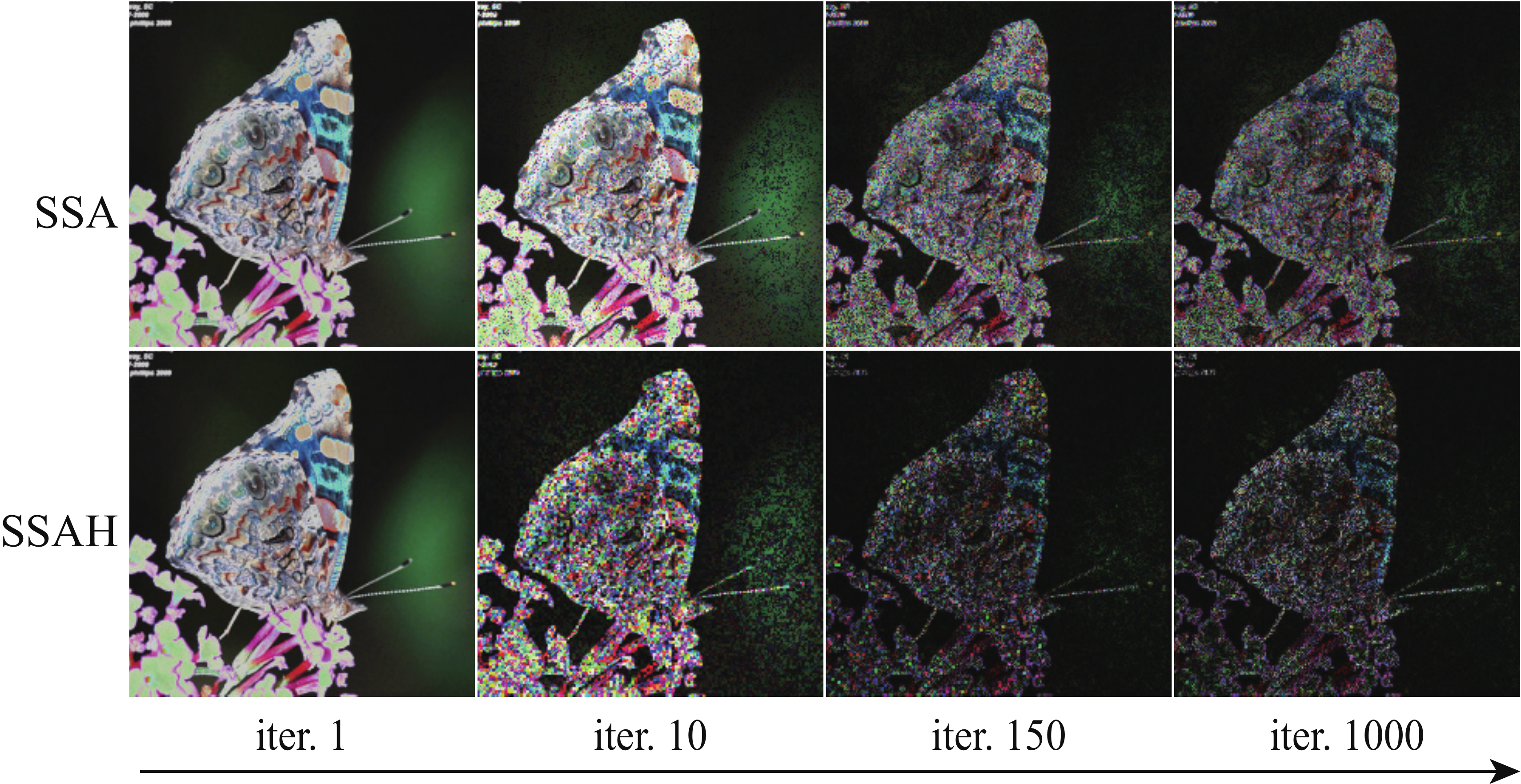} 
  \caption{Normalized perturbations generated by SSA and SSAH in different iterations.}
  \label{fig:ablation_perturbations}
\end{figure}
To study the performance of the proposed low-frequency constraint, normalized perturbations by SSA and SSAH in different iterations are visualized in \cref{fig:ablation_perturbations}. 
This figure shows that the perturbations generated by SSA are prone to distribute on object foreground as well as some smooth background regions, whereas the perturbations by SSAH gradually appear in edges or complex textures.

\begin{table}[htb]
  \centering
  \small
  \begin{tabular}{cccccc}
  \toprule
  \multicolumn{2}{c}{Attack}  &$\ell_2$ $\downarrow$  &$\ell_{\infty}$ $\downarrow$ &FID $\downarrow$ &LF $\downarrow$\\
  \midrule
  \multicolumn{2}{c}{SSA w/o SPW}  &3.12  &\textbf{0.01}  &5.83 &1.40\\
  \multicolumn{2}{c}{SSA}  &2.34 &\textbf{0.01}    &4.63 &1.05\\
  \multicolumn{2}{c}{SSAH}  &\textbf{1.81}  &\textbf{0.01}  &\textbf{3.90} &\textbf{0.06}\\
  \bottomrule
  \end{tabular}    
  \caption{Ablation study of the proposed attack on ImageNet-1K based on different modules, \emph{i.e.}, self-paced weighting  (SPW), low-frequency constraint $\mathcal{D}_{lf}$. w/o SPW means deleting the self-spaced weighting.}
  \label{tab:ex_alabtion}
  \vspace{-0.1in}
\end{table}
To quantify the contribution of each component in SSAH, we conduct an ablation study in \cref{tab:ex_alabtion}.
The results in the 1st and 2nd rows of \cref{tab:ex_alabtion} show SSA, with the adjustment of SPW, significantly reduces FID by a margin of 1.20 compared to the variant without SPW.
The results in the 2nd and 3rd rows show that the proposed low-frequency constraint largely improves LF, \emph{i.e.}, from 1.05 to 0.06.

\section{Conclusion}
We propose a novel framework, SSAH, for adversarial attack. 
It aims to perturb images by attacking their semantic similarity in representation space. 
Such an approach of attacking images in the feature space works well in a variety of settings.
The proposed framework, in particular, could be used in more general and practical black-box settings, such as generating transferable adversarial examples across architectures and datasets and misleading actual online models on various platforms while retaining high imperceptibility.
It is more common in practice to attack in the open set scenario.
For this reason, developing more effective algorithms in this scenario is worthwhile.
Furthermore, the low-frequency constraint is introduced to limit adversarial perturbations within high-frequency components.
Extensive experiments show that such constrained perturbations improve imperceptibility, particularly in smooth regions.

\section{Acknowledgement}
The work was supported by the National Natural Science Foundation of China under grants no. 61602315, 91959108, 
the Science and Technology Project of Guangdong Province under grant no. 2020A1515010707,  
the Science and Technology Innovation Commission of Shenzhen under grant no. JCYJ20190808165203670.

%%%%%%%%% REFERENCES
\clearpage
{\small
\bibliographystyle{ieee_fullname}
\bibliography{egbib}

\begin{thebibliography}{10}\itemsep=-1pt

\bibitem{carlini2017towards}
Nicholas Carlini and David Wagner.
\newblock Towards evaluating the robustness of neural networks.
\newblock In {\em IEEE Symposium on Security and Privacy ({S\&P})}, pages
  39--57, 2017.

\bibitem{chen2018ead}
Pin-Yu Chen, Yash Sharma, Huan Zhang, Jinfeng Yi, and Cho-Jui Hsieh.
\newblock Ead: Elastic-net attacks to deep neural networks via adversarial
  examples.
\newblock In {\em Proceedings of the AAAI Conference on Artificial Intelligence
  ({AAAI})}, 2018.

\bibitem{croce2020reliable}
Francesco Croce and Matthias Hein.
\newblock Reliable evaluation of adversarial robustness with an ensemble of
  diverse parameter-free attacks.
\newblock In {\em International Conference on Machine Learning ({ICML})}, pages
  2206--2216, 2020.

\bibitem{dong2018boosting}
Yinpeng Dong, Fangzhou Liao, Tianyu Pang, Hang Su, Jun Zhu, Xiaolin Hu, and
  Jianguo Li.
\newblock Boosting adversarial attacks with momentum.
\newblock In {\em Proceedings of the IEEE Conference on Computer Vision and
  Pattern Recognition ({CVPR})}, pages 9185--9193, 2018.

\bibitem{duan2021advdrop}
Ranjie Duan, Yuefeng Chen, Dantong Niu, Yun Yang, AK Qin, and Yuan He.
\newblock Advdrop: Adversarial attack to dnns by dropping information.
\newblock In {\em Proceedings of the IEEE International Conference on Computer
  Vision ({ICCV})}, pages 7506--7515, 2021.

\bibitem{duan2017sar}
Yiping Duan, Fang Liu, Licheng Jiao, Peng Zhao, and Lu Zhang.
\newblock Sar image segmentation based on convolutional-wavelet neural network
  and markov random field.
\newblock {\em Pattern Recognition}, 64:255--267, 2017.

\bibitem{feng2020adversarial}
Yan Feng, Bin Chen, Tao Dai, and Shu-Tao Xia.
\newblock Adversarial attack on deep product quantization network for image
  retrieval.
\newblock In {\em Proceedings of the AAAI Conference on Artificial Intelligence
  ({AAAI})}, volume~34, pages 10786--10793, 2020.

\bibitem{goodfellow2014explaining}
Ian~J Goodfellow, Jonathon Shlens, and Christian Szegedy.
\newblock Explaining and harnessing adversarial examples.
\newblock In {\em International Conference on Learning Representations
  ({ICLR})}, 2015.

\bibitem{GuoRCM18}
Chuan Guo, Mayank Rana, Moustapha Ciss{\'{e}}, and Laurens van~der Maaten.
\newblock Countering adversarial images using input transformations.
\newblock In {\em International Conference on Learning Representations
  ({ICLR})}, 2018.

\bibitem{hameed2021perceptually}
Muhammad~Zaid Hameed and Andras Gyorgy.
\newblock Perceptually constrained adversarial attacks.
\newblock {\em arXiv preprint arXiv:2102.07140}, 2021.

\bibitem{he2016deep}
Kaiming He, Xiangyu Zhang, Shaoqing Ren, and Jian Sun.
\newblock Deep residual learning for image recognition.
\newblock In {\em Proceedings of the IEEE Conference on Computer Vision and
  Pattern Recognition ({CVPR})}, pages 770--778, 2016.

\bibitem{heusel2017gans}
Martin Heusel, Hubert Ramsauer, Thomas Unterthiner, Bernhard Nessler, and Sepp
  Hochreiter.
\newblock Gans trained by a two time-scale update rule converge to a local nash
  equilibrium.
\newblock {\em Advances in Neural Information Processing Systems ({NIPS})}, 30,
  2017.

\bibitem{dense-net}
Gao Huang, Zhuang Liu, Laurens Van Der~Maaten, and Kilian~Q Weinberger.
\newblock Densely connected convolutional networks.
\newblock In {\em Proceedings of the IEEE Conference on Computer Vision and
  Pattern Recognition ({CVPR})}, pages 2261--2269, 2017.

\bibitem{inkawhich2019feature}
Nathan Inkawhich, Wei Wen, Hai~Helen Li, and Yiran Chen.
\newblock Feature space perturbations yield more transferable adversarial
  examples.
\newblock In {\em Proceedings of the IEEE Conference on Computer Vision and
  Pattern Recognition ({CVPR})}, pages 7066--7074, 2019.

\bibitem{jang2017objective}
Uyeong Jang, Xi Wu, and Somesh Jha.
\newblock Objective metrics and gradient descent algorithms for adversarial
  examples in machine learning.
\newblock In {\em Annual Computer Security Applications Conference ({ACSAC})},
  pages 262--277, 2017.

\bibitem{krizhevsky2009learning}
Alex Krizhevsky and Geoffrey Hinton.
\newblock Learning multiple layers of features from tiny images.
\newblock Tech report, Department of Computer Science, University of Toronto,
  Toronto, ON, Canada, 2009.

\bibitem{krizhevsky2012imagenet}
Alex Krizhevsky, Ilya Sutskever, and Geoffrey~E Hinton.
\newblock Imagenet classification with deep convolutional neural networks.
\newblock {\em Advances in Neural Information Processing Systems ({NIPS})},
  25:1097--1105, 2012.

\bibitem{kurakin2016adversarial}
Alexey Kurakin, Ian Goodfellow, and Samy Bengio.
\newblock Adversarial examples in the physical world.
\newblock In {\em International Conference on Learning Representations ({ICLR})
  Workshops}, 2017.

\bibitem{laidlaw2020perceptual}
Cassidy Laidlaw, Sahil Singla, and Soheil Feizi.
\newblock Perceptual adversarial robustness: Defense against unseen threat
  models.
\newblock In {\em International Conference on Learning Representations
  ({ICLR})}, 2021.

\bibitem{li2019universal}
Jie Li, Rongrong Ji, Hong Liu, Xiaopeng Hong, Yue Gao, and Qi Tian.
\newblock Universal perturbation attack against image retrieval.
\newblock In {\em Proceedings of the IEEE International Conference on Computer
  Vision ({ICCV})}, pages 4899--4908, 2019.

\bibitem{li2020wavelet}
Qiufu Li, Linlin Shen, Sheng Guo, and Zhihui Lai.
\newblock Wavelet integrated cnns for noise-robust image classification.
\newblock In {\em Proceedings of the IEEE Conference on Computer Vision and
  Pattern Recognition ({CVPR})}, pages 7245--7254, 2020.

\bibitem{liu2018multi}
Pengju Liu, Hongzhi Zhang, Kai Zhang, Liang Lin, and Wangmeng Zuo.
\newblock Multi-level wavelet-cnn for image restoration.
\newblock In {\em Proceedings of the IEEE Conference on Computer Vision and
  Pattern Recognition ({CVPR}) workshops}, pages 773--782, 2018.

\bibitem{madry2017towards}
Aleksander Madry, Aleksandar Makelov, Ludwig Schmidt, Dimitris Tsipras, and
  Adrian Vladu.
\newblock Towards deep learning models resistant to adversarial attacks.
\newblock In {\em International Conference on Learning Representations
  ({ICLR})}, 2018.

\bibitem{maiya2021frequency}
Shishira~R Maiya, Max Ehrlich, Vatsal Agarwal, Ser-Nam Lim, Tom Goldstein, and
  Abhinav Shrivastava.
\newblock A frequency perspective of adversarial robustness.
\newblock {\em arXiv preprint arXiv:2111.00861}, 2021.

\bibitem{mopuri2018generalizable}
Konda~Reddy Mopuri, Aditya Ganeshan, and R~Venkatesh Babu.
\newblock Generalizable data-free objective for crafting universal adversarial
  perturbations.
\newblock {\em IEEE Transactions on Pattern Analysis and Machine Intelligence},
  41(10):2452--2465, 2018.

\bibitem{panareda2017open}
Pau Panareda~Busto and Juergen Gall.
\newblock Open set domain adaptation.
\newblock In {\em Proceedings of the IEEE International Conference on Computer
  Vision ({ICCV})}, pages 754--763, 2017.

\bibitem{papernot2016limitations}
Nicolas Papernot, Patrick McDaniel, Somesh Jha, Matt Fredrikson, Z~Berkay
  Celik, and Ananthram Swami.
\newblock The limitations of deep learning in adversarial settings.
\newblock In {\em 2016 IEEE European Symposium on Security and Privacy
  ({EuroS\&P})}, pages 372--387, 2016.

\bibitem{rony2019decoupling}
J{\'e}r{\^o}me Rony, Luiz~G Hafemann, Luiz~S Oliveira, Ismail {Ben Ayed},
  Robert Sabourin, and Eric Granger.
\newblock Decoupling direction and norm for efficient gradient-based l2
  adversarial attacks and defenses.
\newblock In {\em Proceedings of the IEEE Conference on Computer Vision and
  Pattern Recognition ({CVPR})}, 2019.

\bibitem{russakovsky2015imagenet}
Olga Russakovsky, Jia Deng, Hao Su, Jonathan Krause, Sanjeev Satheesh, Sean Ma,
  Zhiheng Huang, Andrej Karpathy, Aditya Khosla, Michael Bernstein, et~al.
\newblock Imagenet large scale visual recognition challenge.
\newblock {\em International Journal of Computer Vision}, 115(3):211--252,
  2015.

\bibitem{sabour2016adversarial}
Sara Sabour, Yanshuai Cao, Fartash Faghri, and David~J Fleet.
\newblock Adversarial manipulation of deep representations.
\newblock In {\em International Conference on Learning Representations
  ({ICLR})}, 2016.

\bibitem{Sharif_2018_CVPR_Workshops}
Mahmood Sharif, Lujo Bauer, and Michael~K. Reiter.
\newblock On the suitability of lp-norms for creating and preventing
  adversarial examples.
\newblock In {\em Proceedings of the IEEE Conference on Computer Vision and
  Pattern Recognition ({CVPR}) Workshops}, June 2018.

\bibitem{sharma2019effectiveness}
Yash Sharma, Gavin~Weiguang Ding, and Marcus~A Brubaker.
\newblock On the effectiveness of low frequency perturbations.
\newblock In {\em Proceedings of the AAAI Conference on Artificial Intelligence
  ({AAAI})}, pages 3389--3396, 2019.

\bibitem{simonyan2014very}
Karen Simonyan and Andrew Zisserman.
\newblock Very deep convolutional networks for large-scale image recognition.
\newblock In {\em International Conference on Learning Representations
  ({ICLR})}, 2015.

\bibitem{sun2020circle}
Yifan Sun, Changmao Cheng, Yuhan Zhang, Chi Zhang, Liang Zheng, Zhongdao Wang,
  and Yichen Wei.
\newblock Circle loss: A unified perspective of pair similarity optimization.
\newblock In {\em Proceedings of the IEEE Conference on Computer Vision and
  Pattern Recognition ({CVPR})}, pages 6398--6407, 2020.

\bibitem{szegedy2016rethinking}
Christian Szegedy, Vincent Vanhoucke, Sergey Ioffe, Jon Shlens, and Zbigniew
  Wojna.
\newblock Rethinking the inception architecture for computer vision.
\newblock In {\em Proceedings of the IEEE Conference on Computer Vision and
  Pattern Recognition ({CVPR})}, pages 2818--2826, 2016.

\bibitem{szegedy2014intriguing}
Christian {Szegedy}, Wojciech {Zaremba}, Ilya {Sutskever}, Joan {Bruna},
  Dumitru {Erhan}, Ian {Goodfellow}, and Rob {Fergus}.
\newblock Intriguing properties of neural networks.
\newblock In {\em International Conference on Learning Representations
  ({ICLR})}, 2014.

\bibitem{TramerKPGBM18}
Florian Tram{\`{e}}r, Alexey Kurakin, Nicolas Papernot, Ian~J. Goodfellow, Dan
  Boneh, and Patrick~D. McDaniel.
\newblock Ensemble adversarial training: Attacks and defenses.
\newblock In {\em International Conference on Learning Representations
  ({ICLR})}, 2018.

\bibitem{wang2020transferable}
Hongjun Wang, Guangrun Wang, Ya Li, Dongyu Zhang, and Liang Lin.
\newblock Transferable, controllable, and inconspicuous adversarial attacks on
  person re-identification with deep mis-ranking.
\newblock In {\em Proceedings of the IEEE Conference on Computer Vision and
  Pattern Recognition ({CVPR})}, pages 342--351, 2020.

\bibitem{wang2020high}
Haohan Wang, Xindi Wu, Zeyi Huang, and Eric~P Xing.
\newblock High-frequency component helps explain the generalization of
  convolutional neural networks.
\newblock In {\em Proceedings of the IEEE Conference on Computer Vision and
  Pattern Recognition ({CVPR})}, pages 8684--8694, 2020.

\bibitem{wang2019advpattern}
Zhibo Wang, Siyan Zheng, Mengkai Song, Qian Wang, Alireza Rahimpour, and
  Hairong Qi.
\newblock advpattern: Physical-world attacks on deep person re-identification
  via adversarially transformable patterns.
\newblock In {\em Proceedings of the IEEE International Conference on Computer
  Vision ({ICCV})}, pages 8341--8350, 2019.

\bibitem{wong2018provable}
Eric Wong and Zico Kolter.
\newblock Provable defenses against adversarial examples via the convex outer
  adversarial polytope.
\newblock In {\em International Conference on Machine Learning ({ICML})}, pages
  5286--5295, 2018.

\bibitem{Xu0Q18}
Weilin Xu, David Evans, and Yanjun Qi.
\newblock Feature squeezing: Detecting adversarial examples in deep neural
  networks.
\newblock In {\em Annual Network and Distributed System Security Symposium
  ({NDSS})}, 2018.

\bibitem{NEURIPS2019_b05b57f6}
Dong Yin, Raphael Gontijo~Lopes, Jon Shlens, Ekin~Dogus Cubuk, and Justin
  Gilmer.
\newblock A fourier perspective on model robustness in computer vision.
\newblock In {\em Advances in Neural Information Processing Systems ({NIPS})},
  volume~32, 2019.

\bibitem{BMVC2016_87}
Sergey Zagoruyko and Nikos Komodakis.
\newblock Wide residual networks.
\newblock In {\em Proceedings of the British Machine Vision Conference
  ({BMVC})}, pages 87.1--87.12, September 2016.

\bibitem{zhang2019defense}
Haichao Zhang and Jianyu Wang.
\newblock Defense against adversarial attacks using feature scattering-based
  adversarial training.
\newblock {\em Advances in Neural Information Processing Systems ({NIPS})},
  32:1831--1841, 2019.

\bibitem{zhang2019theoretically}
Hongyang Zhang, Yaodong Yu, Jiantao Jiao, Eric Xing, Laurent El~Ghaoui, and
  Michael Jordan.
\newblock Theoretically principled trade-off between robustness and accuracy.
\newblock In {\em International Conference on Machine Learning ({ICML})}, pages
  7472--7482, 2019.

\bibitem{zhang2018unreasonable}
Richard Zhang, Phillip Isola, Alexei~A Efros, Eli Shechtman, and Oliver Wang.
\newblock The unreasonable effectiveness of deep features as a perceptual
  metric.
\newblock In {\em Proceedings of the IEEE Conference on Computer Vision and
  Pattern Recognition ({CVPR})}, pages 586--595, 2018.

\bibitem{zhao2020towards}
Zhengyu Zhao, Zhuoran Liu, and Martha Larson.
\newblock Towards large yet imperceptible adversarial image perturbations with
  perceptual color distance.
\newblock In {\em Proceedings of the IEEE Conference on Computer Vision and
  Pattern Recognition ({CVPR})}, pages 1039--1048, 2020.

\end{thebibliography}
}
\clearpage

\appendix
\section{Methodology for Targeted Attack}
As mentioned above, we define the attack in the targeted scenario as:
\renewcommand{\theequation}{4}
\begin{equation}
        \vx_i^{adv}  = \argmin_{\vx'_i} [ s'_{i,i} - s'_{i,t} ]_+,
\end{equation}
where $t$ denotes the index of the target image in the minibatch, $s'_{i,i} = \text{sim}(f(\vx'_i), f(\vx_i))$ and $s_{i,t} = \text{sim}(f(\vx'_i), f(\vx_t))$ are similarity scores.
It aims to encourage the adversarial example $\vx'_i$ to be close to the target $\vx_t$ in terms of the feature representation. 

The self-weighting scheme that is specific to this targeted attack is defined as follows:
\renewcommand{\theequation}{11}
\begin{equation}
        \begin{aligned}
            \vx_i^{adv}  &= \argmin_{\vx'_i} \mathcal{L}_{SSA}(\vx_i,\vx'_i) \\
             &= \argmin_{\vx'_i} [ \alpha_{i} s'_{i,i} - \beta_{i} s'_{i,t} ]_+.
        \end{aligned}
        \label{eq:ad_targeted_alpha}
\end{equation}
As for the weighting factors $\alpha_{i}$ and $\beta_{i}$, we set them as:
\renewcommand{\theequation}{12}
\begin{equation}\label{eq:targeted_scale}
    % \footnotesize{
           \left\{
           \begin{aligned}
           \alpha_{i}&=[s'_{i,i} - m]_+,\\
           \beta_{i}&=[1 + m -s'_{i,t}]_+.\\
           \end{aligned}
       \right.
    %    }
\end{equation}
The targeted attack follows a similar attack design and weight factor setting as the untargeted attack.
The only difference is that SSA in the targeted scenario does not need the selection of the most dissimilar example in the minibatch but requires a certain example of the target category.
\section{Implementation details}
For compared attack approaches, the parameters $\kappa$ and $c$ in C$\&$W \cite{carlini2017towards} are set to 40 and 0.1, respectively; the parameters $\alpha_l$ and $\alpha_c$ in PerC-AL \cite{zhao2020towards} are initialized to 1 and 0.5 and gradually reduced to 0.01 and 0.05, respectively, with cosine annealing.
For evaluating the robustness of attacks against defense approaches or the attack success rates of attacks against online models, the $\ell_{\infty}$ bound of 8/255 is used for perturbation generation.

\section{Additional Experimental Results}
\subsection{Targeted Attack in the White-box Setting}
\renewcommand\thetable{6}
\begin{table*}
  \centering
  \small
  \scalebox{1}{
  \begin{tabular}{clccccccc}
  \toprule
  Dataset & Attack & Iteration &RunTime (s) $\downarrow$ & ASR (\%) $\uparrow$ &$\ell_2$ $\downarrow$ &$\ell_{\infty}$ $\downarrow$   & FID $\downarrow$ & LF $\downarrow$\\
  \midrule[1pt] 
  \multirow{8}{*}{CIFAR-10}
      & BIM \cite{kurakin2016adversarial} &10 &33 &99.94 &0.67 &\textbf{0.03} &13.36  &0.18\\
      & PGD \cite{madry2017towards} & 10 &38 &99.94 &1.19 &\textbf{0.03} &27.23 &0.31\\
      & MIM \cite{dong2018boosting}  & 10 &35 &99.92  &1.88 &\textbf{0.03} &26.67  & 0.47\\
      & AdvDrop \cite{duan2021advdrop} & 150 &389 &99.11 &1.13 & 0.08 &16.52 &0.42\\
      & C$\&$W $\ell_2$ \cite{carlini2017towards} &1000 &978 &100  &\textbf{0.45} &0.07 &10.34 &0.13\\
      &\cellcolor{gray!30}SSA (ours) &\cellcolor{gray!30}150 &\cellcolor{gray!30}176 &\cellcolor{gray!30}99.90  &\cellcolor{gray!30}0.55&\cellcolor{gray!30}0.04 &\cellcolor{gray!30}6.13 &\cellcolor{gray!30}0.15\\
      &\cellcolor{gray!30}SSAH (ours) &\cellcolor{gray!30}150 &\cellcolor{gray!30}178 &\cellcolor{gray!30}99.92 &\cellcolor{gray!30}0.48& \cellcolor{gray!30}0.04&\cellcolor{gray!30}\textbf{5.14} &\cellcolor{gray!30}\textbf{0.07}\\
  \midrule[0.75pt]
  \multirow{8}{*}{CIFAR-100}
  & BIM \cite{kurakin2016adversarial} & 10 &32 & 99.41 &\textbf{0.74} & \textbf{0.03} &13.59 & 0.27\\
  & PGD \cite{madry2017towards} & 10 & 36 & 99.34 &1.22 & \textbf{0.03} & 25.64 &0.39\\
  & MIM \cite{dong2018boosting}  & 10 & 32 & 99.11  &1.84 & \textbf{0.03} & 25.49 & 0.64\\
  & AdvDrop \cite{duan2021advdrop} & 150 &308 & 97.70 &1.09 & 0.08 &15.56 &0.43\\
  & C$\&$W $\ell_2$ \cite{carlini2017towards} &1000 &743 &99.99  &0.94 & 0.09 &17.59 & 0.64\\
  &\cellcolor{gray!30}SSA (ours) &\cellcolor{gray!30}150 &\cellcolor{gray!30}134 &\cellcolor{gray!30}99.15  &\cellcolor{gray!30}1.27 &\cellcolor{gray!30}0.08 &\cellcolor{gray!30}10.57 &\cellcolor{gray!30}0.52\\
  &\cellcolor{gray!30}SSAH (ours) &\cellcolor{gray!30}150 &\cellcolor{gray!30}138 &\cellcolor{gray!30}99.01 &\cellcolor{gray!30}0.99 &\cellcolor{gray!30}0.07&\cellcolor{gray!30}\textbf{8.88} &\cellcolor{gray!30}\textbf{0.07}\\

  \midrule[0.75pt]
  \multirow{8}{*}{ImageNet-1K}
  & BIM \cite{kurakin2016adversarial} & 10 & 2166 &98.31 &25.18 & 0.03 &39.61 &10.44\\
  & PGD \cite{madry2017towards} & 10 &2973 &98.65 &53.84 & 0.03 &37.21 &16.87\\
  & MIM \cite{dong2018boosting}  &10 &2358 &99.98  &92.86 &0.03 &81.62 & 39.93\\
  & AdvDrop \cite{duan2021advdrop} &150 &46968 & 99.76 &14.95 & 0.06 & 11.28 &5.67\\
  & C$\&$W $\ell_2$ \cite{carlini2017towards} &1000 & $>100000$ & 97.83  &\textbf{1.85} & 0.04 &12.93 &0.84\\
  & \cellcolor{gray!30}SSA (ours) &\cellcolor{gray!30}200 &\cellcolor{gray!30}31742 &\cellcolor{gray!30}98.64  &\cellcolor{gray!30}4.31 &\cellcolor{gray!30}\textbf{0.02} &\cellcolor{gray!30}8.64 & \cellcolor{gray!30}1.94\\
  & \cellcolor{gray!30}SSAH (ours) &\cellcolor{gray!30}200 &\cellcolor{gray!30}30050 &\cellcolor{gray!30}98.06 &\cellcolor{gray!30}3.38 &\cellcolor{gray!30}\textbf{0.02} &\cellcolor{gray!30}\textbf{6.42} &\cellcolor{gray!30}\textbf{0.47}\\
  \bottomrule[1pt]
  \end{tabular}} 
  \caption{Results of the attack success rate (ASR) and four metrics related with perceptual similarity by seven attack approaches in the targeted scenario. The best results are marked in bold.}
  \label{ex:targeted_attack}
\end{table*}
In this section, we investigate the attack performance of SSAH in the targeted scenario in Tab.~\ref{ex:targeted_attack}.
It shows that our SSAH remains effective at generating imperceptible perturbations for the targeted scenario.

\subsection{Parameter Sensitivity Analyses}
\renewcommand\thefigure{8}
\begin{figure}[htb]
    \centering
    \includegraphics[width=0.83\columnwidth]{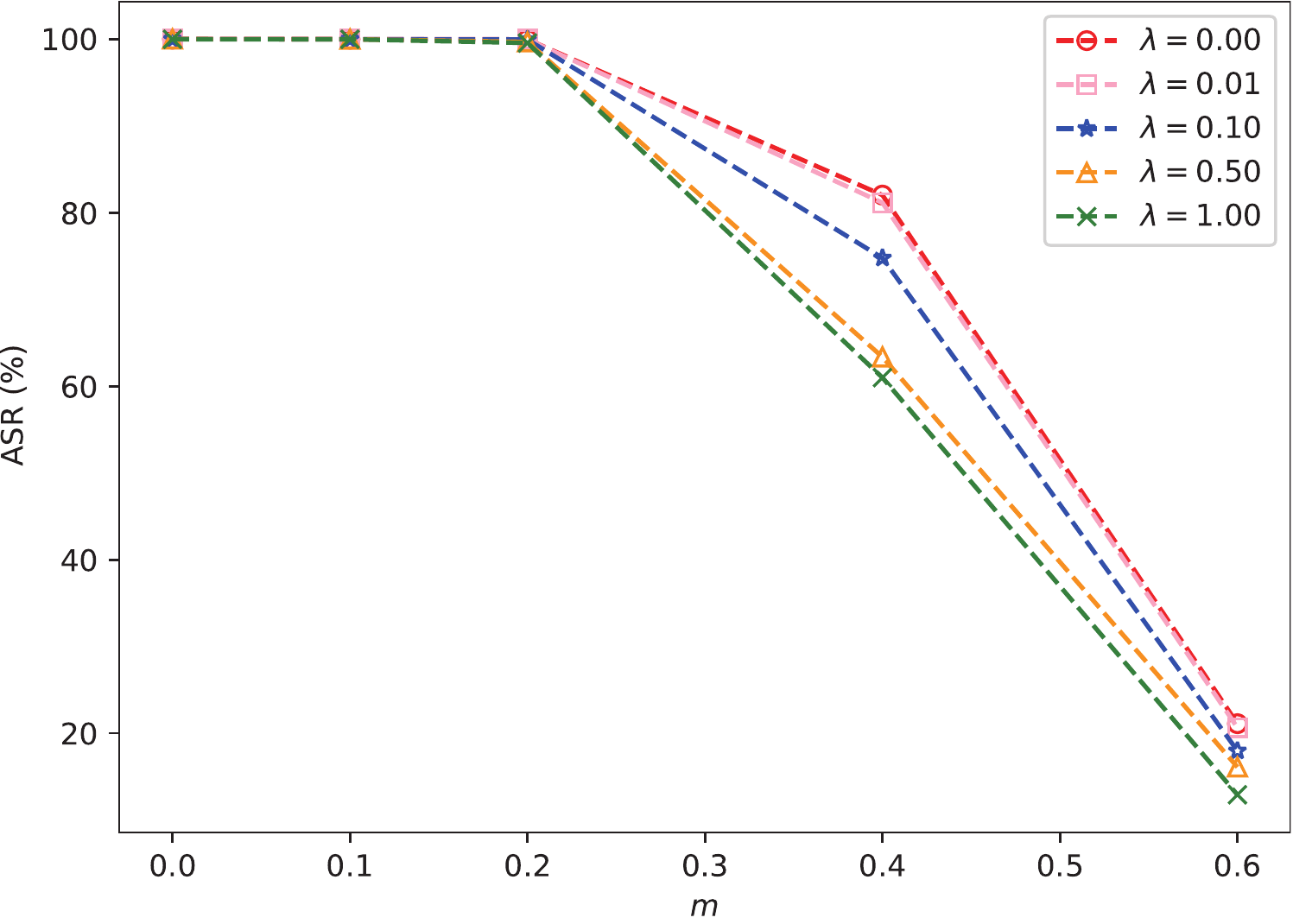} % Reduce the figure size so that it is slightly narrower than the column. Don't use precise values for figure width.This setup will avoid overfull boxes.
    \caption{Sensitive analyses of $\lambda$ and $m$ in terms of attack success rate (ASR).}
    \label{fig:parameter_sensitivity_asr}
\end{figure}
\renewcommand\thefigure{9}
\begin{figure}[htb]
    \centering
    \includegraphics[width=0.83\columnwidth]{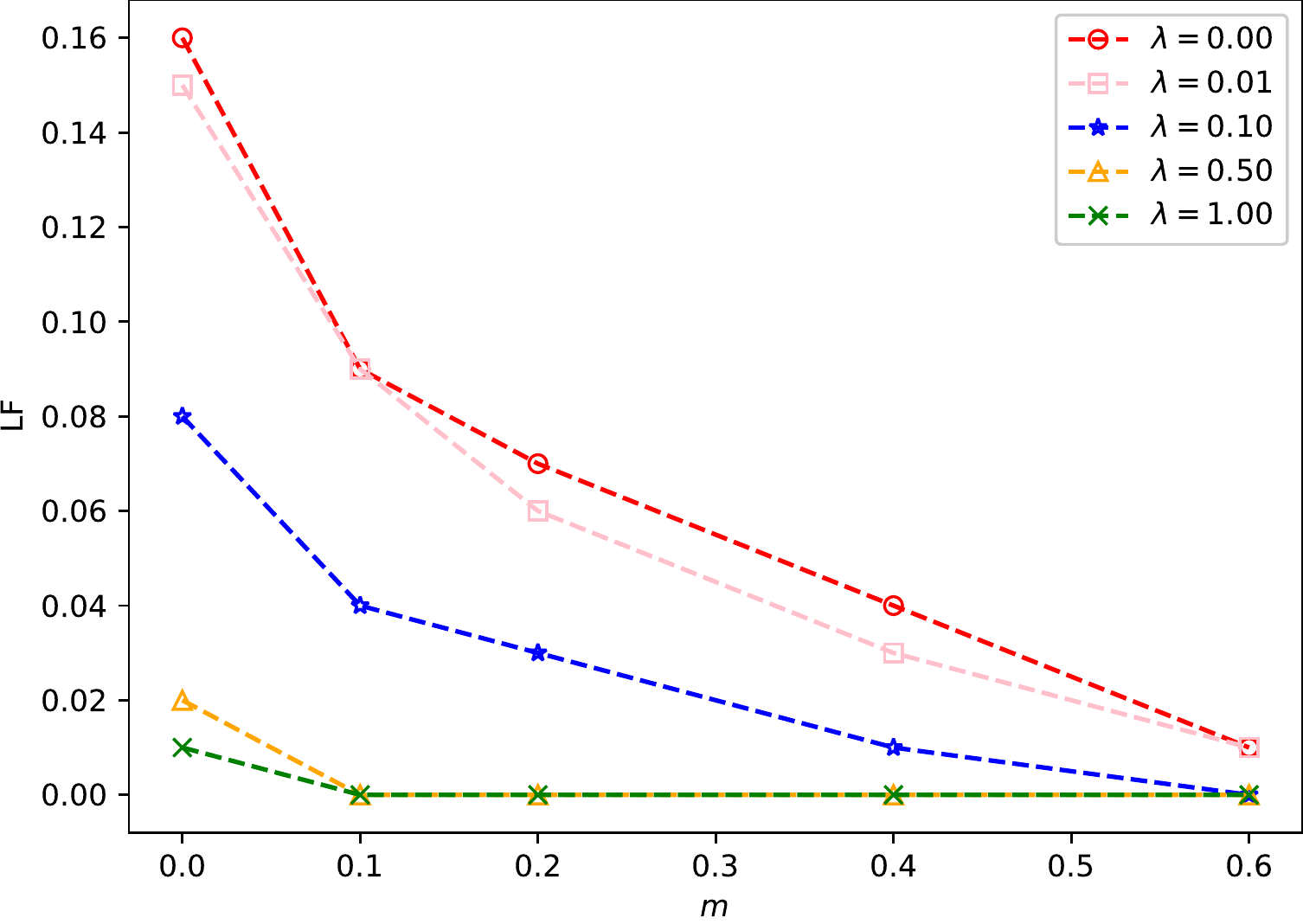} % Reduce the figure size so that it is slightly narrower than the column. Don't use precise values for figure width.This setup will avoid overfull boxes.
    \caption{Sensitive analyses of $\lambda$ and $m$ in terms of LF.}
    \label{fig:parameter_sensitivity_lf}
    \vspace{-0.1in}
\end{figure}
\renewcommand\thefigure{10}
\begin{figure*}[htb]
    \centering
    \includegraphics[width=1.82\columnwidth]{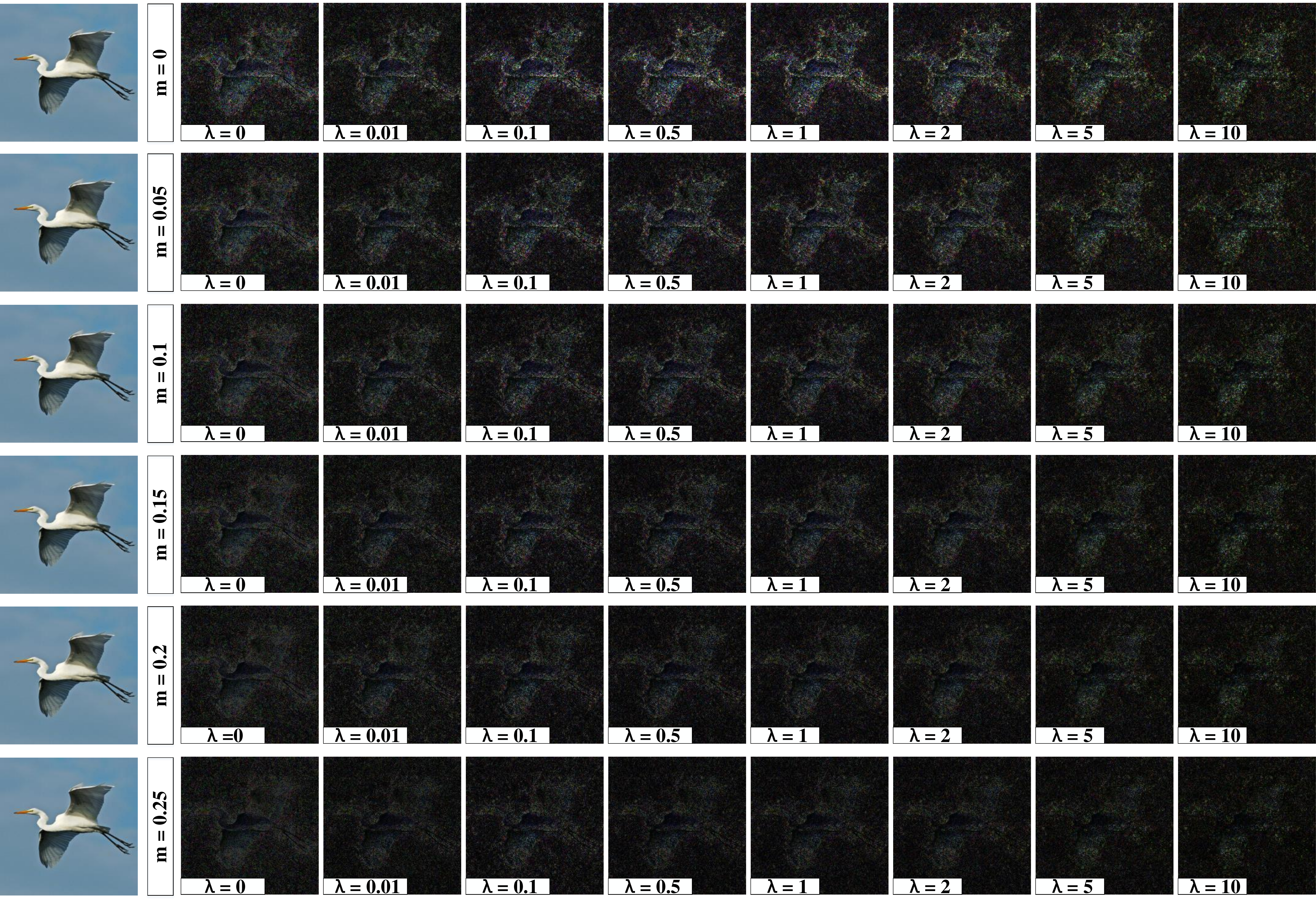} % Reduce the figure size so that it is slightly narrower than the column. Don't use precise values for figure width.This setup will avoid overfull boxes.
    \caption{Adversarial perturbations to an ImageNet-1K image using different hyperparameters of $m$ and $\lambda$ in SSAH.}
    \label{fig:parameter_sensitivity_examples}
    \vspace{-0.1in}
\end{figure*}
There are two hyperparameters in SSAH, \emph{i.e.}, the margin $m$ in Eq.~(6) for adjusting self-paced weighting and $\lambda$ in Eq.~(10) for weighing the low-frequency constraint.
The sensitivity analyses of these two hyperparameters are performed on CIFAR-10 and ImageNet-1K.
The quantitative results on CIFAR-10 are presented in Figs.~\ref{fig:parameter_sensitivity_asr} and ~\ref{fig:parameter_sensitivity_lf}, and the visualization results on ImageNet-1K are presented in Fig.~\ref{fig:parameter_sensitivity_examples}.

Fig.~\ref{fig:parameter_sensitivity_asr} shows that SSAH becomes sensitive to $\lambda$ when the margin $m$ is large (\emph{e.g.}, $m \ge 0.2$). 
Taking the case of $m = 0.4$ for example, along with increasing weighting $\lambda$, the performance of SSAH decreases fast from 82.02\% when $\lambda = 0.00$ to 60.98\% when $\lambda = 1.00$.
The reason is that we actually impose a weak attack strength on images when $m$ is a large value.
In this condition, a large $\lambda$ results in a strong constraint on our attack objective in SSAH, leading to a low attack success rate.

Based on the above analysis, it can be concluded that an appropriate selection of $m$ and $\lambda$ is necessary.
The sensitivity analyses in Fig.~\ref{fig:parameter_sensitivity_asr} and Fig.~\ref{fig:parameter_sensitivity_lf} show that SSAH can achieve relatively stable and satisfying performances when $m \in [0.0,0.2]$ and $\lambda \in [0.1,1.0]$.

\subsection{Batch Size}
Tab.~\ref{ex:batch_size} reports the results with a batch size from 32 to 10000. 
Our attack works reasonably well over this wide range of batch sizes.
The results are similarly good when the batch size is from 32 to 10000, and the differences are at the level of random variations.
\begin{table}[htb]
    \renewcommand\thetable{7}
   \centering
       \begin{tabular}{ccccc}
           \toprule
       Batch Size    &  ASR $\uparrow$   &$\ell_2$ $\downarrow$  &$\ell_{\infty}$ $\downarrow$  &LF $\downarrow$             \\    \midrule
       32    & 99.94          & \textbf{0.25} & \textbf{0.02} & \textbf{0.02} \\
       64    & 99.90          & \textbf{0.25} & \textbf{0.02}  & \textbf{0.02} \\
       128   & 99.91          & \textbf{0.25} & \textbf{0.02}  & 0.03          \\
       256   & \textbf{99.95} & \textbf{0.25} & \textbf{0.02}  & 0.03          \\
       512   & 99.94          & \textbf{0.25} & \textbf{0.02}  & 0.03          \\
       1024  & 99.91          & \textbf{0.25} & \textbf{0.02}  & 0.03          \\
       2048  & 99.89          & 0.26          & \textbf{0.02}  & 0.03          \\
       4096  & 99.90          & 0.26          & \textbf{0.02} & 0.03          \\
       10000 & 99.94          & 0.26          & \textbf{0.02}  & 0.03         \\\bottomrule
       \end{tabular}
       \caption{Effect of batch sizes (CIFAR-10 evaluation attack success rate and four metrics related with perceptual similarity).}
       \label{ex:batch_size}
       \vspace{-0.2in}
\end{table}

\subsection{Additional Evaluation of Imperceptibility}
In this section, we evaluate perturbation imperceptibility in terms of perceptual colour difference ($\overline{C_2} $).
This metric is used in PerC-AL \cite{zhao2020towards} to measure human colour perception.

Tab. \ref{ex:untargeted_c2} shows that our SSAH without the constrant of $\ell_2$ ($\overline{C_2}$) distance still achieves competitive performances in terms of $\ell_2$ ($\overline{C_2}$) .
It further demonstrates the superiority of our attack in perturbation imperceptibility.
\begin{table}[htb]
  \renewcommand\thetable{8}
  \centering
  \begin{tabular}{cccccc}
      \toprule
      Attack& Iter. &ASR $\uparrow$ &$\ell_2$ $\downarrow$ &LF $\downarrow$ &$\overline{C_2}$ $\downarrow$   \\
      \midrule
      C$\&$W &1000  &99.27 &\textbf{1.51} &0.67  &152.51 \\
      PerC-AL&1000  &98.78  &4.35 &1.59  &\textbf{90.62} \\\midrule
      SSAH (ours) &200  &98.01  &1.81  &\textbf{0.06}  &124.32  \\\bottomrule
      \end{tabular}
      \caption{Results of the attack success rate (ASR) and three metrics related with perceptual similarity by three attack approaches on ImageNet-1K in the untargeted scenario.}
      \label{ex:untargeted_c2}
      \vspace{-0.1in}
  \end{table}

\subsection{Additional Robustness Evaluations}
\renewcommand\thefigure{11}
\begin{figure}[htb]
    \centering
    \includegraphics[width=0.80\columnwidth]{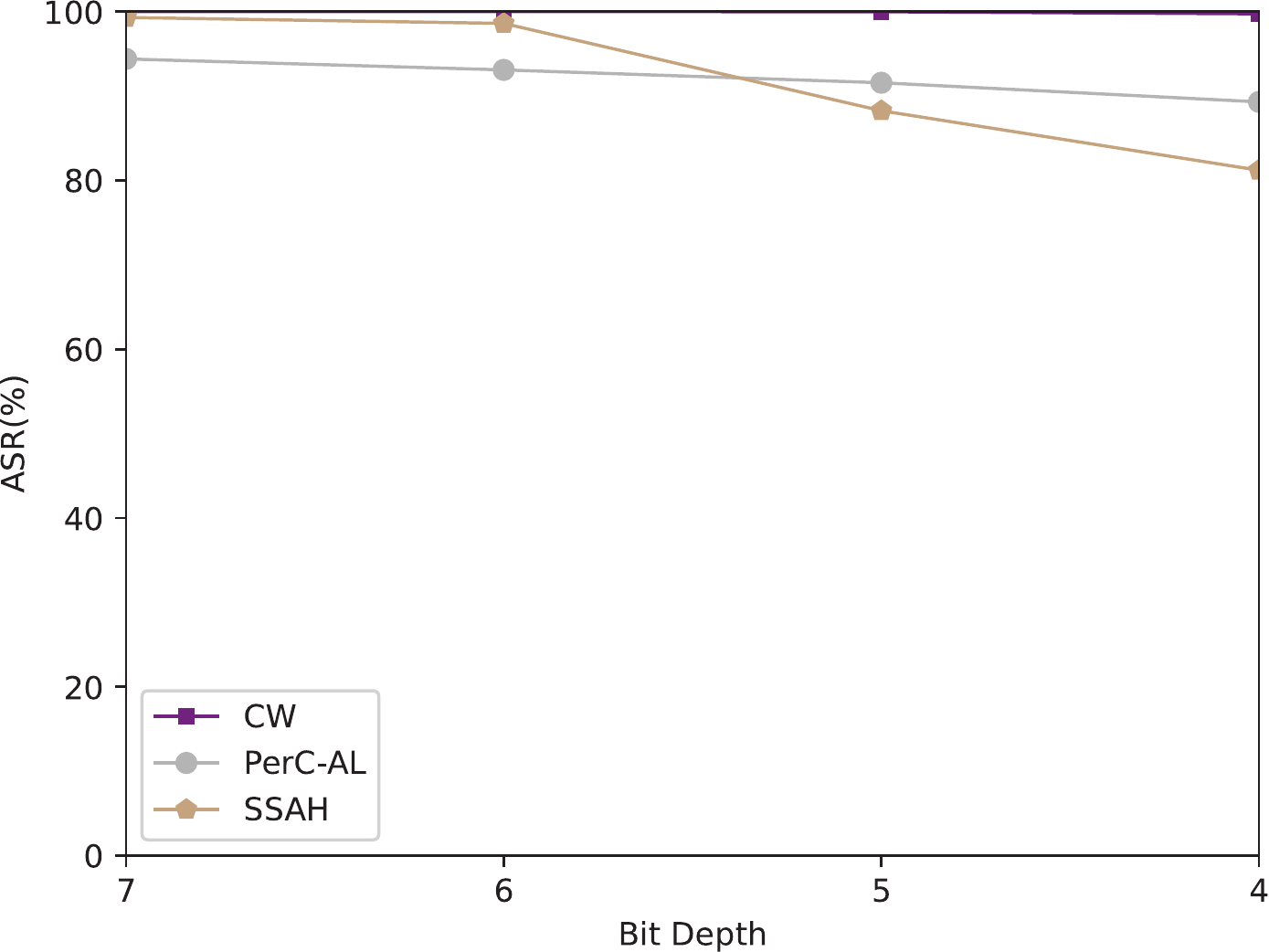} % Reduce the figure size so that it is slightly narrower than the column. Don't use precise values for figure width.This setup will avoid overfull boxes.
    \caption{Evaluation of robustness of adversarial examples generated by three attack approaches against bit-depth reduction on CIFAR-10.}
    \label{fig:appendix_adversarial_examples_against_bit_depth-reduction}
    \vspace{-0.1in}
\end{figure}
To evaluate the performance of our attack against image transformation-based defense, we test the robustness of the adversarial examples against bit-depth reduction \cite{GuoRCM18,Xu0Q18}.
Fig.~\ref{fig:appendix_adversarial_examples_against_bit_depth-reduction} shows our imperceptible attack on high-frequency components is still robust against this image transformation-based defense (i,e., bit-depth reduction).

\subsection{Analysis of SPW}
\renewcommand\thefigure{12}
\begin{figure}[htb]
    \centering
    \includegraphics[width=0.9\columnwidth]{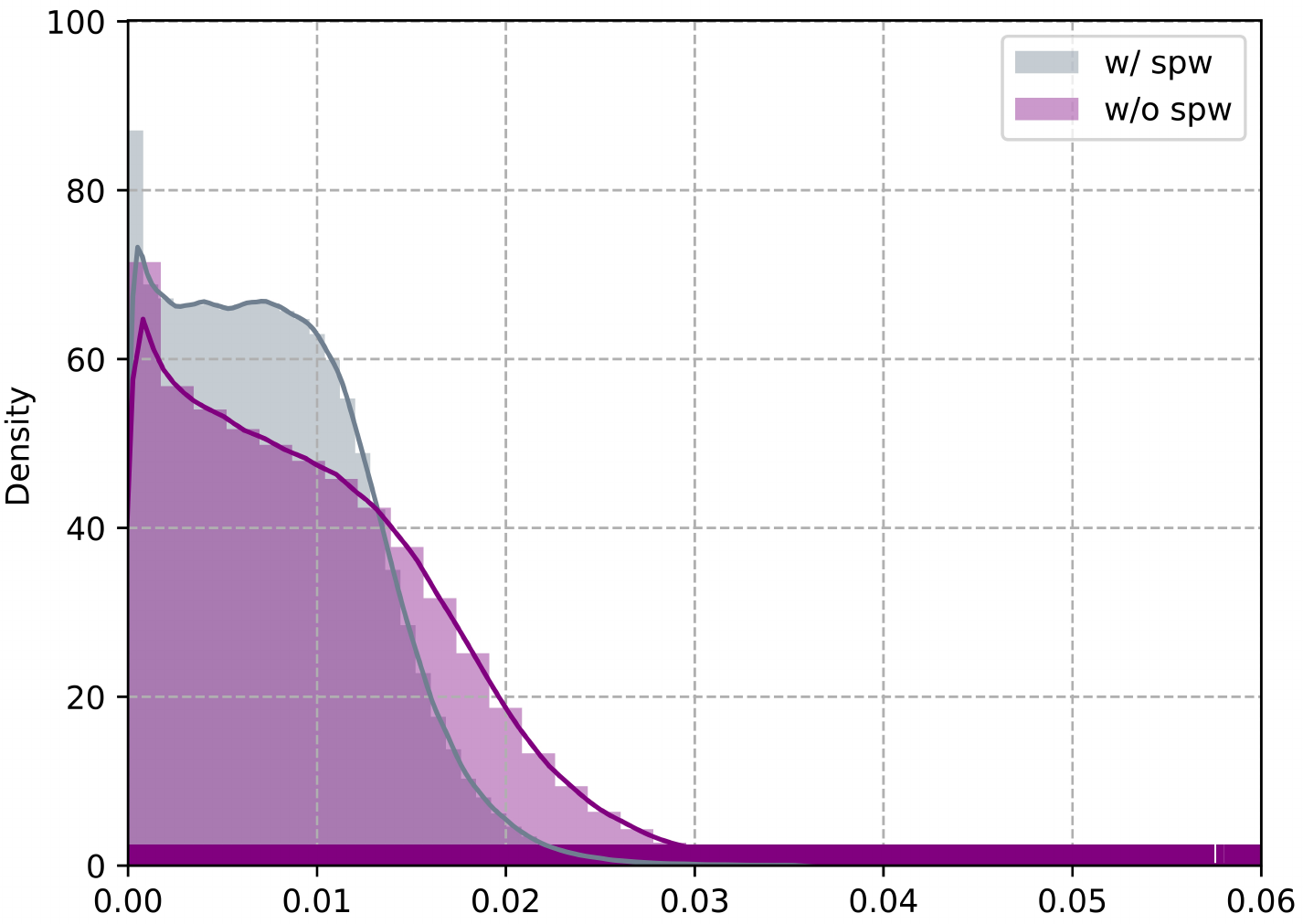} % Reduce the figure size so that it is slightly narrower than the column. Don't use precise values for figure width.This setup will avoid overfull boxes.
    \caption{Distributions of perturbation intensities by SSAH with self-paced weighting (w/ SPW) and without self-paced weighting (w/o SPW) on a subset of 1024 random samples from CIFAR-10.}
    \label{fig:ablation_perturbations_distribution}
    \vspace{-0.2in}
\end{figure}
To further measure the effect of applying the self-paced weighting in SSAH, we randomly sample images from the testing set of CIFAR-10, and present the distributions of perturbation intensities generated by SSAH with and without self-paced weighting in Fig.~\ref{fig:ablation_perturbations_distribution}.
This figure shows that the perturbation intensities generated by SSAH with SPW appear to be smaller than those by SSAH without SPW. 
It indicates that the weighting scheme can well reduce the redundant perturbations caused by over-optimization.

\subsection{Additional Visualization Results}
In this section, we present more visualization results, including adversarial examples and perturbations generated in the white-box setting
 (\emph{i.e.}, Fig.~\ref{fig:appendix_adversarial_examples_cifar10}, Fig.~\ref{fig:appendix_adversarial_examples_cifar100}, Fig.~\ref{fig:appendix_adversarial_examples} and Fig.~\ref{fig:appendix_adversarial_examples_perturbations}) and transferable adversarial examples across architectures and datasets (\emph{i.e.}, Fig.~\ref{fig:appendix_adversarial_examples_cross_dataset_architecture}).

\renewcommand\thefigure{13}
\begin{figure}[htb]
    \centering
    \includegraphics[width=0.98\columnwidth]{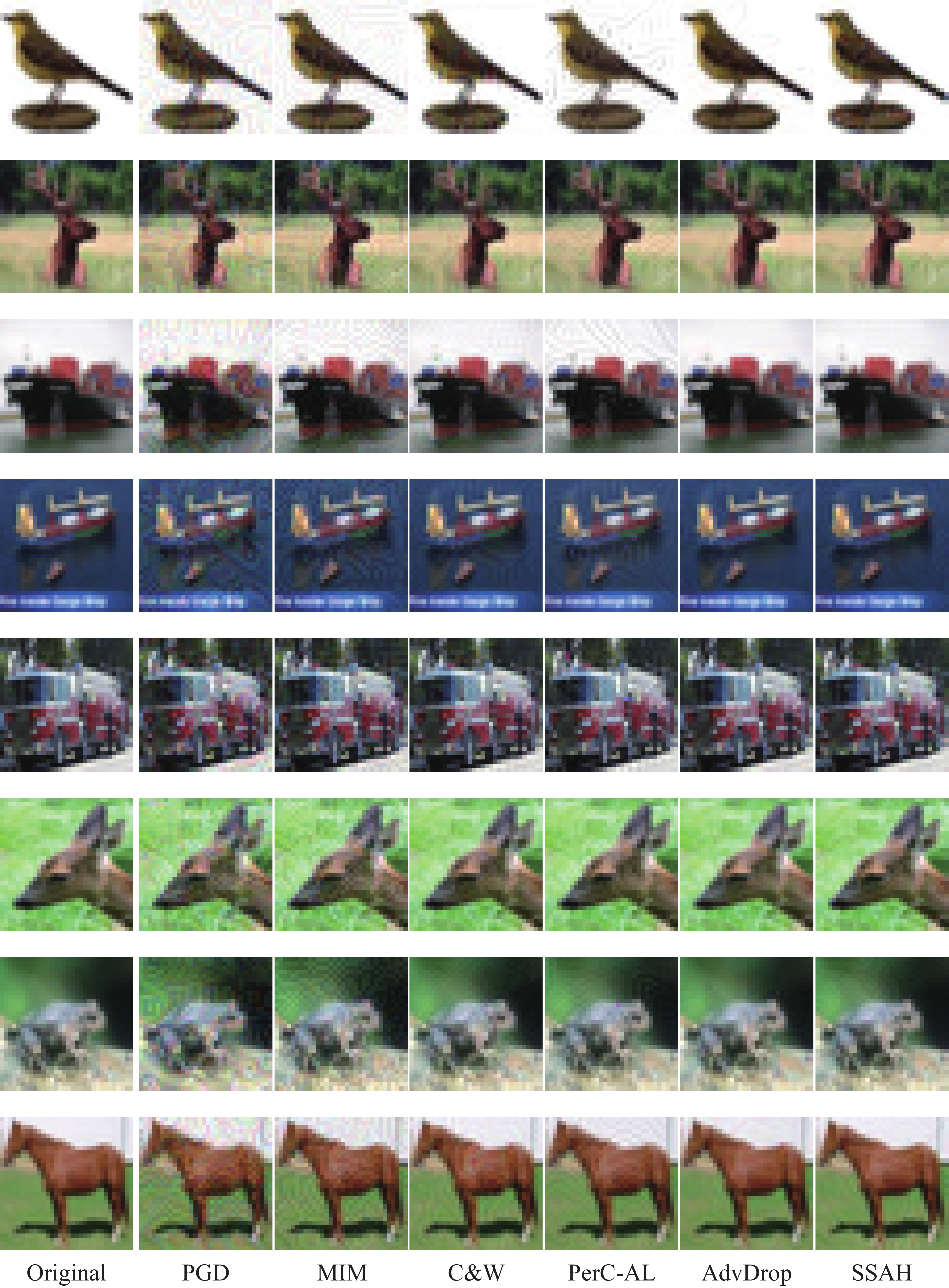} % Reduce the figure size so that it is slightly narrower than the column. Don't use precise values for figure width.This setup will avoid overfull boxes.
    \caption{Adversarial examples generated by six different attack approaches for CIFAR-10.}
    \label{fig:appendix_adversarial_examples_cifar10}
    \vspace{-0.1in}
\end{figure}

\renewcommand\thefigure{14}
\begin{figure}[htb]
    \centering
    \includegraphics[width=0.98\columnwidth]{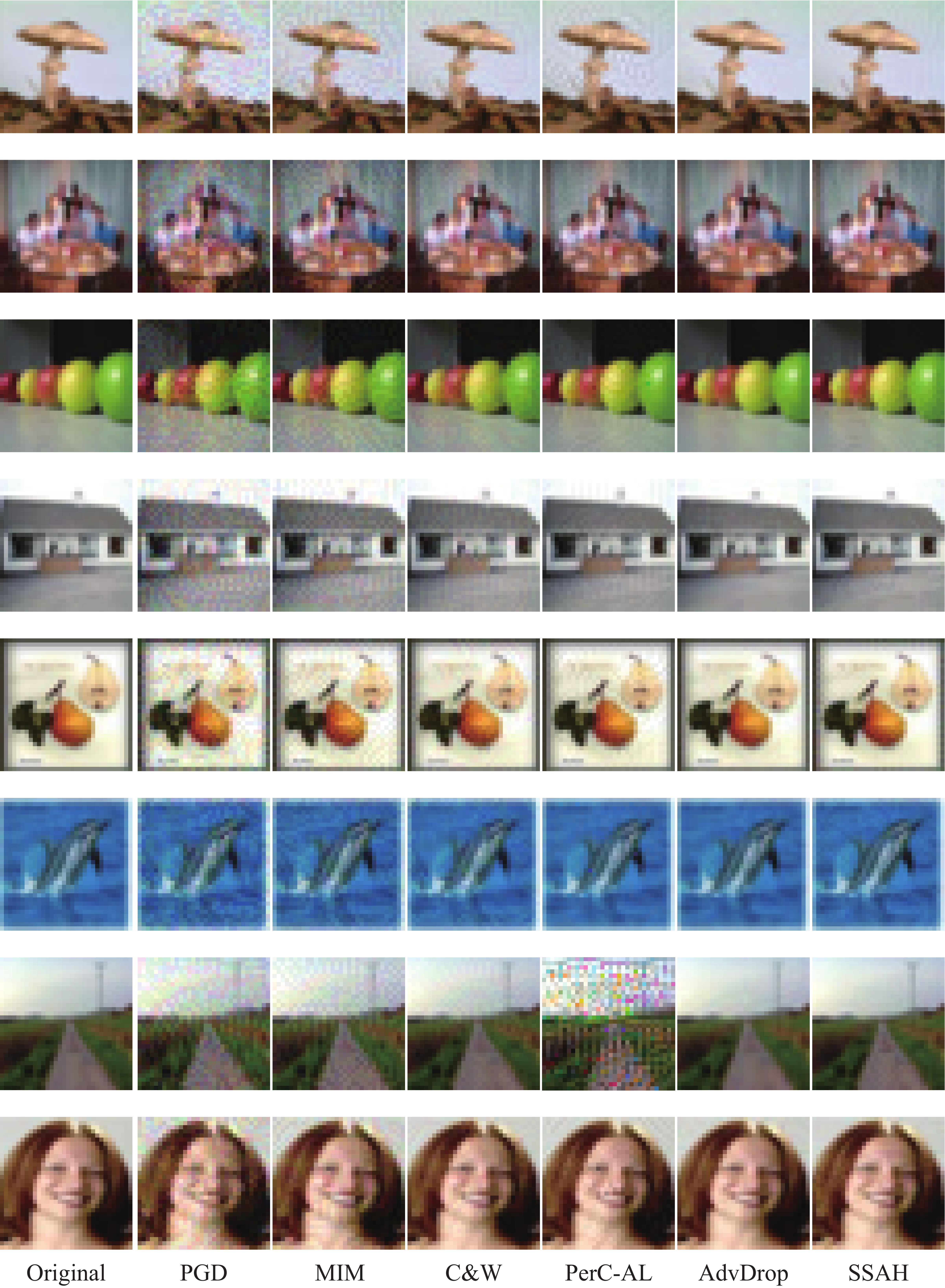} % Reduce the figure size so that it is slightly narrower than the column. Don't use precise values for figure width.This setup will avoid overfull boxes.
    \caption{Adversarial examples generated by six different attack approaches for CIFAR-100.}
    \label{fig:appendix_adversarial_examples_cifar100}
    \vspace{-0.1in}
\end{figure}
\clearpage
\renewcommand\thefigure{15}
\begin{figure*}[htb]
    \centering
    \includegraphics[width=2\columnwidth]{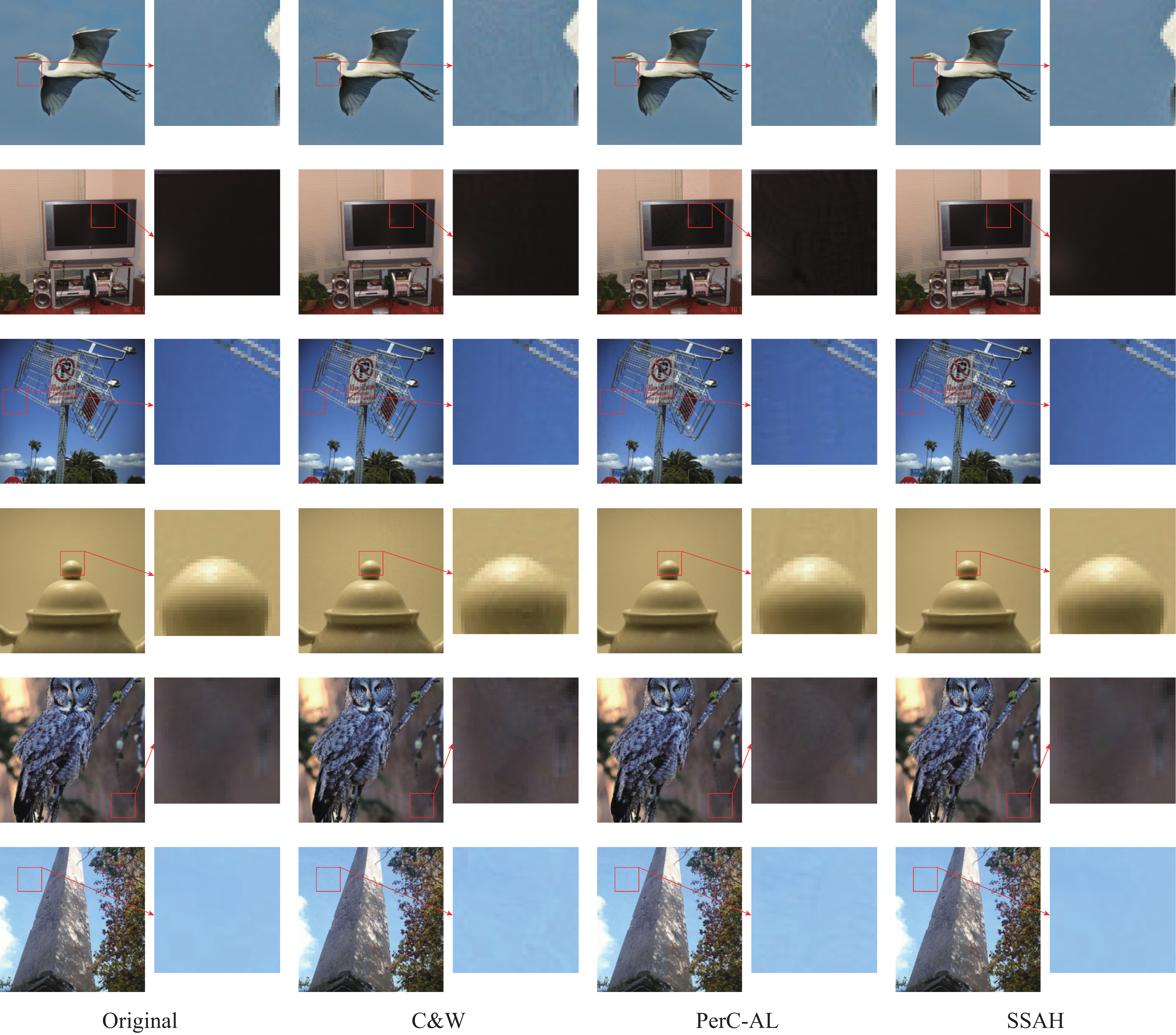} % Reduce the figure size so that it is slightly narrower than the column. Don't use precise values for figure width.This setup will avoid overfull boxes.
    \caption{Adversarial examples generated by three different attack approaches for ImageNet-1K.}
    \label{fig:appendix_adversarial_examples}
\end{figure*}

\renewcommand\thefigure{16}
\begin{figure*}[h!]
    \centering
    \includegraphics[width=2\columnwidth]{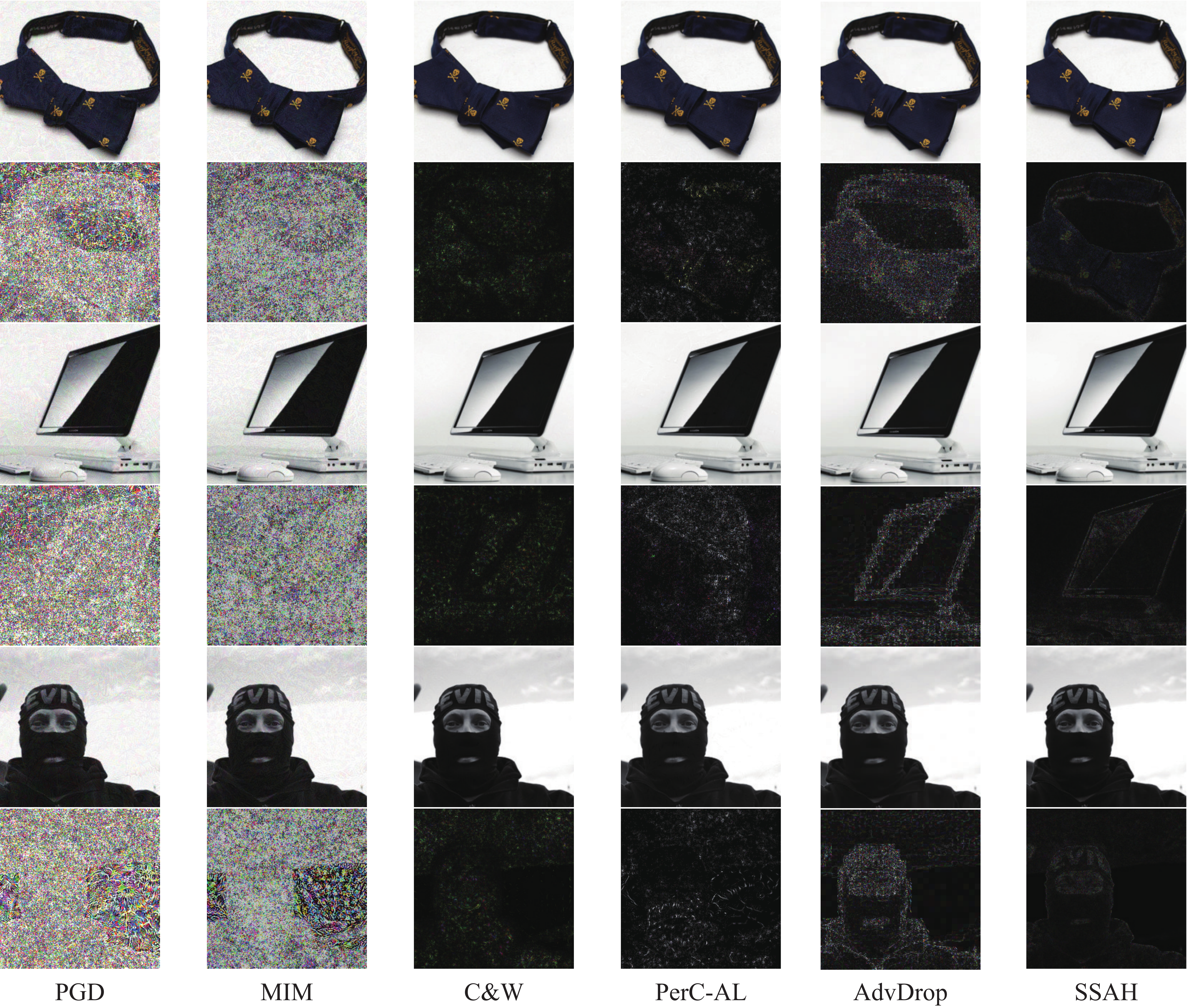} % Reduce the figure size so that it is slightly narrower than the column. Don't use precise values for figure width.This setup will avoid overfull boxes.
    \caption{Adversarial examples and their corresponding perturbations generated by six different attack approaches for ImageNet-1K. The 1st, 3rd and 5th rows are adversarial examples, while the 2nd, 4th and 6th rows are their perturbations.}
    \label{fig:appendix_adversarial_examples_perturbations}
\end{figure*}

\renewcommand\thefigure{17}
\begin{figure*}[h!]
    \centering
    \includegraphics[width=2\columnwidth]{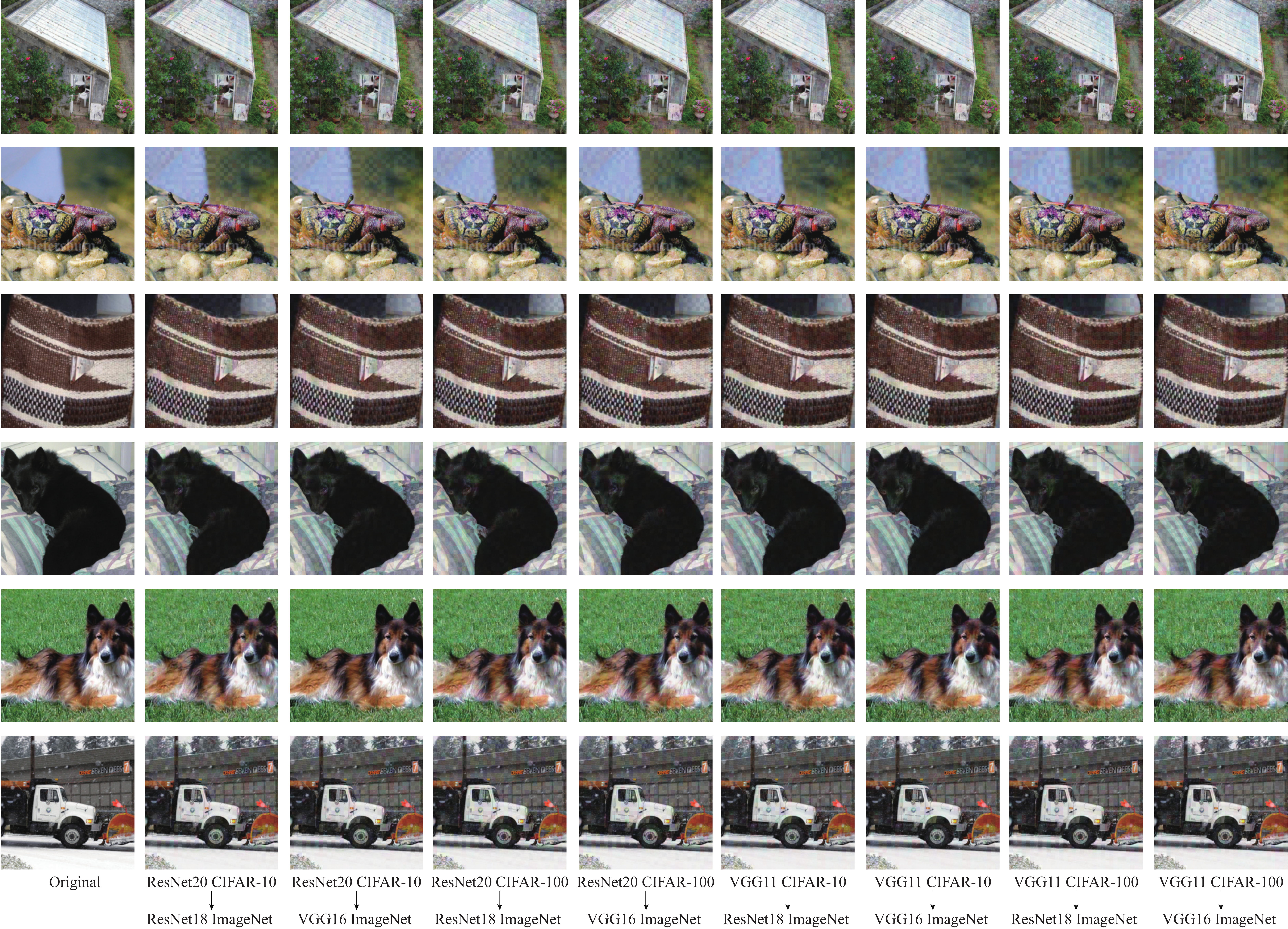} % Reduce the figure size so that it is slightly narrower than the column. Don't use precise values for figure width.This setup will avoid overfull boxes.
    \caption{Adversarial examples generated by SSAH based on a surrogate model trained on the source domain to a target model trained on the target domain. The original examples are selected from the validation set of ImageNet-1K. A B $\rightarrow$ C D denotes that model A trained on dataset B is used to craft perturbations to fool model C trained on dataset D.}
    \label{fig:appendix_adversarial_examples_cross_dataset_architecture}
\end{figure*}

\clearpage
\section{Image Samples in Transferable Attack}
In the experiment of attacking online models, we randomly sample 200 images from the ImageNet-1K validation set.
For reproducibility, we list the names of these used images as follows:

\noindent n02108089/val\_00016416, 
n01917289/val\_00047927, 
n03877845/val\_00026991,
n02125311/val\_00024128,
n03661043/val\_00018698,
n02085620/val\_00031154,
n03691459/val\_00023671,
n03888257/val\_00025121,
n02276258/val\_00032773,
n01704323/val\_00038815,
n02342885/val\_00027868,
n01632458/val\_00031521,
n03770439/val\_00013597,
n02231487/val\_00009943,
n03478589/val\_00040225,
n02111277/val\_00046591,
n02840245/val\_00031888,
n04398044/val\_00041709,
n07892512/val\_00017146,
n01694178/val\_00025511,
n03873416/val\_00047900,
n02114367/val\_00033884,
n02092002/val\_00022857,
n03976467/val\_00033699,
n02342885/val\_00041792,
n02457408/val\_00041441,
n03773504/val\_00022433,
n03930313/val\_00002877,
n09193705/val\_00038734,
n02804414/val\_00008402,
n03124170/val\_00026251,
n01806143/val\_00023973,
n01818515/val\_00021663,
n03376595/val\_00040795,
n02226429/val\_00045770,
n02655020/val\_00008184,
n02484975/val\_00045387,
n03478589/val\_00000035,
n02951585/val\_00023091,
n01692333/val\_00033244,
n02281406/val\_00046852,
n02389026/val\_00014369,
n04310018/val\_00038429,
n03956157/val\_00038501,
n04200800/val\_00018851,
n01968897/val\_00030526,
n01608432/val\_00043085,
n03444034/val\_00026796,
n02342885/val\_00047769,
n02814533/val\_00005978,
n01843383/val\_00037244,
n02422106/val\_00035337,
n02963159/val\_00030024,
n13052670/val\_00039616,
n01829413/val\_00036044,
n02823750/val\_00041752,
n03602883/val\_00039677,
n12985857/val\_00038482,
n04005630/val\_00004526,
n04487394/val\_00033068,
n03127747/val\_00011868,
n02701002/val\_00028205,
n03124170/val\_00033776,
n03355925/val\_00006767,
n03042490/val\_00025402,
n02787622/val\_00019653,
n03837869/val\_00036456,
n04523525/val\_00019004,
n04409515/val\_00014503,
n13052670/val\_00033352,
n02074367/val\_00023213,
n03075370/val\_00021941,
n02971356/val\_00023953,
n03126707/val\_00043871,
n02641379/val\_00008847,
n01440764/val\_00017699,
n03457902/val\_00049086,
n03180011/val\_00043506,
n01980166/val\_00005836,
n04392985/val\_00003756,
n07747607/val\_00005888,
n04417672/val\_00046924,
n02165105/val\_00030936,
n03290653/val\_00015959,
n03933933/val\_00030150,
n03709823/val\_00001860,
n02028035/val\_00010037,
n04118538/val\_00045773,
n02361337/val\_00037480,
n03529860/val\_00022699,
n04204347/val\_00036559,
n02606052/val\_00014777,
n02104365/val\_00035049,
n02494079/val\_00002579,
n01877812/val\_00022693,
n01687978/val\_00029055,
n02107683/val\_00035628,
n02804414/val\_00047642,
n03891332/val\_00036752,
n02113186/val\_00009045,
n02105162/val\_00030963,
n02172182/val\_00045513,
n01560419/val\_00018672,
n02134084/val\_00016340,
n02794156/val\_00006960,
n01484850/val\_00016988,
n03016953/val\_00021529,
n04404412/val\_00049571,
n03888257/val\_00023107,
n02909870/val\_00036628,
n02105641/val\_00030831,
n09468604/val\_00028882,
n02097658/val\_00015119,
n02172182/val\_00031062,
n02106030/val\_00014342,
n02493793/val\_00013076,
n03976467/val\_00032721,
n01924916/val\_00037245,
n01944390/val\_00022381,
n03954731/val\_00005903,
n02883205/val\_00003045,
n04004767/val\_00008345,
n01693334/val\_00022937,
n02443484/val\_00018383,
n04597913/val\_00016472,
n03670208/val\_00010364,
n02009912/val\_00026861,
n02326432/val\_00002413,
n04229816/val\_00047467,
n02971356/val\_00000831,
n04192698/val\_00001173,
n02791270/val\_00044488,
n03595614/val\_00025275,
n03126707/val\_00038038,
n04613696/val\_00014904,
n04235860/val\_00006430,
n01484850/val\_00045730,
n02835271/val\_00033559,
n02128385/val\_00036527,
n02395406/val\_00033668,
n02115913/val\_00026888,
n02930766/val\_00049552,
n02403003/val\_00017987,
n02655020/val\_00005213,
n04049303/val\_00018090,
n04553703/val\_00048784,
n03868863/val\_00011676,
n03769881/val\_00001207,
n04399382/val\_00006985,
n03394916/val\_00003759,
n07760859/val\_00025973,
n04069434/val\_00001854,
n02093647/val\_00000883,
n02690373/val\_00030319,
n01728920/val\_00020274,
n02132136/val\_00006703,
n04141327/val\_00003236,
n03743016/val\_00038715,
n02102480/val\_00004185,
n02974003/val\_00021691,
n03018349/val\_00020100,
n02074367/val\_00030029,
n04149813/val\_00032339,
n03594945/val\_00037128,
n04399382/val\_00039275,
n04372370/val\_00003304,
n03388549/val\_00041363,
n01984695/val\_00002243,
n02489166/val\_00000206,
n02037110/val\_00011196,
n04418357/val\_00028155,
n04251144/val\_00005522,
n02795169/val\_00025956,
n02113799/val\_00016697,
n04461696/val\_00021814,
n09468604/val\_00041897,
n02093991/val\_00004687,
n07802026/val\_00040001,
n03992509/val\_00046414,
n03662601/val\_00023517,
n02110958/val\_00011495,
n03467068/val\_00020240,
n02869837/val\_00015695,
n02127052/val\_00036625,
n11879895/val\_00012320,
n02747177/val\_00002145,
n02966687/val\_00008803,
n02704792/val\_00030917,
n07613480/val\_00000620,
n07930864/val\_00039707,
n04389033/val\_00030415,
n02128385/val\_00024214,
n02391049/val\_00015124,
n01616318/val\_00006250,
n04536866/val\_00024173,
n03467068/val\_00049842,
n03180011/val\_00049615,
n02099429/val\_00040055,
n03291819/val\_00011445, 
n04447861/val\_00038921.
\quad\quad\quad\quad

\end{document}